\RequirePackage{luatex85}

\documentclass[10pt,twocolumn,letterpaper]{article}
\usepackage{iccv}              %

\newcommand{\NAME}{ATLAS}

\newcommand{\jp}[1]{#1}

\usepackage{tabularx}
\usepackage{wrapfig}
\usepackage{multirow}
\usepackage{float}
\usepackage{multicol}
\usepackage{cuted}
\usepackage{stfloats}

\newcolumntype{Y}{>{\centering\arraybackslash}X}
\usepackage{colortbl}  %
\usepackage[dvipsnames]{xcolor}
\usepackage{color}

\newlength\savewidth

\usepackage{amssymb}%
\usepackage{pifont}%
\newcommand{\cmark}{\ding{51}}%
\newcommand{\xmark}{\ding{55}}%
\definecolor{mylightgray}{gray}{0.6}
\makeatletter
\newcommand\notsotiny{\@setfontsize\notsotiny\@vipt\@viipt}
\makeatother

\definecolor{iccvblue}{rgb}{0.21,0.49,0.74}
\usepackage[pagebackref,breaklinks,colorlinks,allcolors=iccvblue]{hyperref}

\def\paperID{12793} %
\def\confName{ICCV}
\def\confYear{2025}

\title{\NAME{}: Decoupling Skeletal and Shape Parameters for Expressive Parametric Human Modeling}

\author{
Jinhyung Park\textsuperscript{1,2} \quad
Javier Romero\textsuperscript{1} \quad
Shunsuke Saito\textsuperscript{1} \quad
Fabian Prada\textsuperscript{1} \quad
Takaaki Shiratori\textsuperscript{1}\\
Yichen Xu\textsuperscript{1} \quad
Federica Bogo\textsuperscript{1} \quad
Shoou-I Yu\textsuperscript{1} \quad
Kris Kitani\textsuperscript{1,2} \quad
Rawal Khirodkar\textsuperscript{1}\\
\textsuperscript{1}Meta \quad
\textsuperscript{2}Carnegie Mellon University \\
\small{\url{https://jindapark.github.io/projects/atlas}}
}

\definecolor{rawal_colour}{RGB}{117,112,178}    

\definecolor{david_colour}{RGB}{0,200,200}    

\begin{document}

\twocolumn[{
    \maketitle
    \begin{figure}[H]
    \hsize=\textwidth
    \centering
    \vspace*{-0.4in}
    \includegraphics[width=\textwidth]{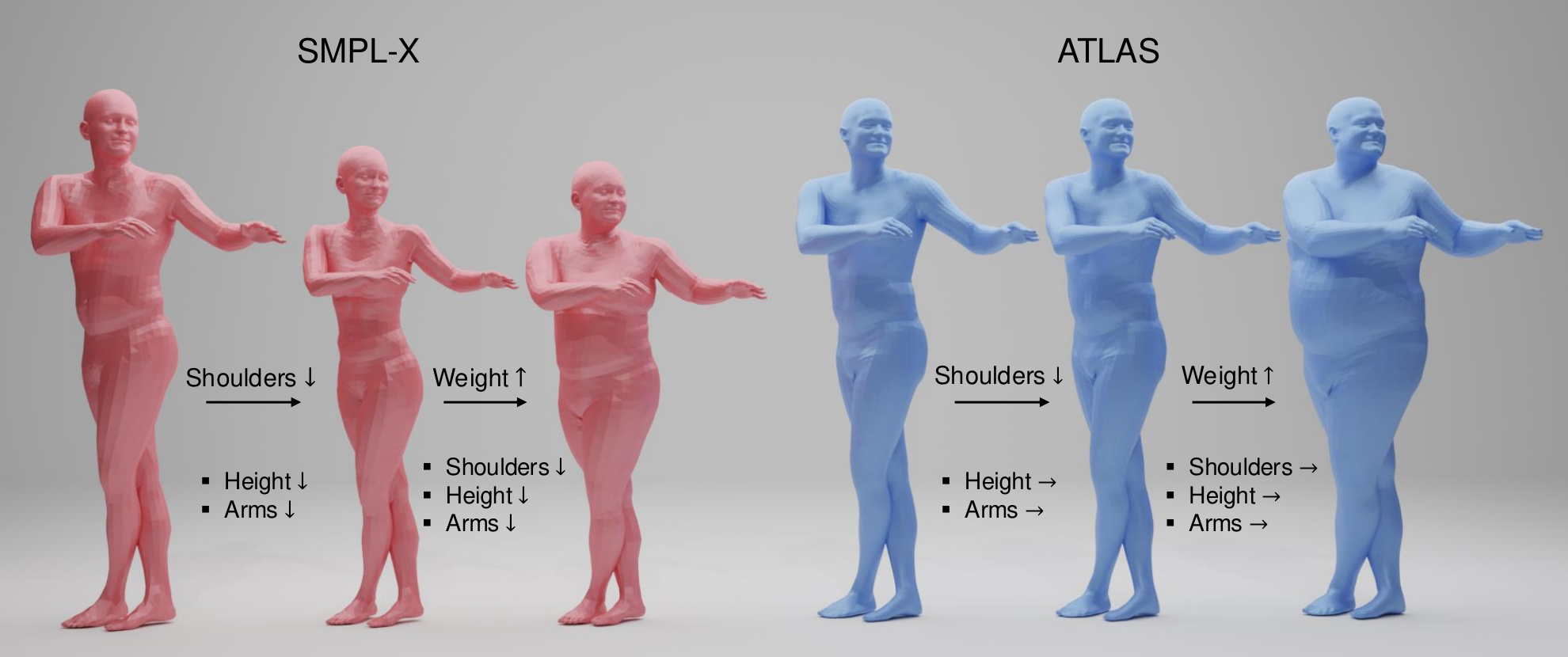}
    \vspace*{-0.2in}
    \caption{\textbf{ATLAS enables precise, decoupled control of skeletal and surface attributes.} Here, we customize a mesh to reduce shoulder width and increase body weight. This level of control is difficult to accomplish in prior work~\cite{SMPL-X:2019} due to undesirable correlations between joints and vertices, \eg adjusting shoulder width affects the entire body and increasing weight reverses the shoulder adjustment. With ATLAS's decoupled skeleton and shape, this customization is a simple two-step deterministic edit.}
    \label{fig:smplx_shoulder_width}
    \vspace*{-0.1in}
    \end{figure}
}]

\maketitle
 \vspace*{-0.1in}
\begin{abstract}
Parametric body models offer expressive 3D representation of humans across a wide range of poses, shapes, and facial expressions, typically derived by learning a basis over registered 3D meshes. However, existing human mesh modeling approaches struggle to capture detailed variations across diverse body poses and shapes, largely due to limited training data diversity and restrictive modeling assumptions. Moreover, the common paradigm first optimizes the external body surface using a linear basis, then regresses internal skeletal joints from surface vertices. This approach introduces problematic dependencies between internal skeleton and outer soft tissue, limiting direct control over body height and bone lengths. To address these issues, we present \NAME{}, a high-fidelity body model learned from $600$k high-resolution scans captured using $240$ synchronized cameras. Unlike previous methods, we explicitly decouple the shape and skeleton bases by grounding our mesh representation in the human skeleton. This decoupling enables enhanced shape expressivity, fine-grained customization of body attributes, and keypoint fitting independent of external soft-tissue characteristics. \NAME{} outperforms existing methods by fitting unseen subjects in diverse poses more accurately, and quantitative evaluations show that our non-linear pose correctives more effectively capture complex poses compared to linear models.
\end{abstract}
\begin{quote}
\vspace{-0.2in}
\centering
``\textit{ATLAS}---a structure bearing human form.''
\end{quote}

\vspace{-0.5em}
\section{Introduction}
\vspace{-0.3em}
\begin{figure}[t]
  \centering
  \includegraphics[width=\linewidth]{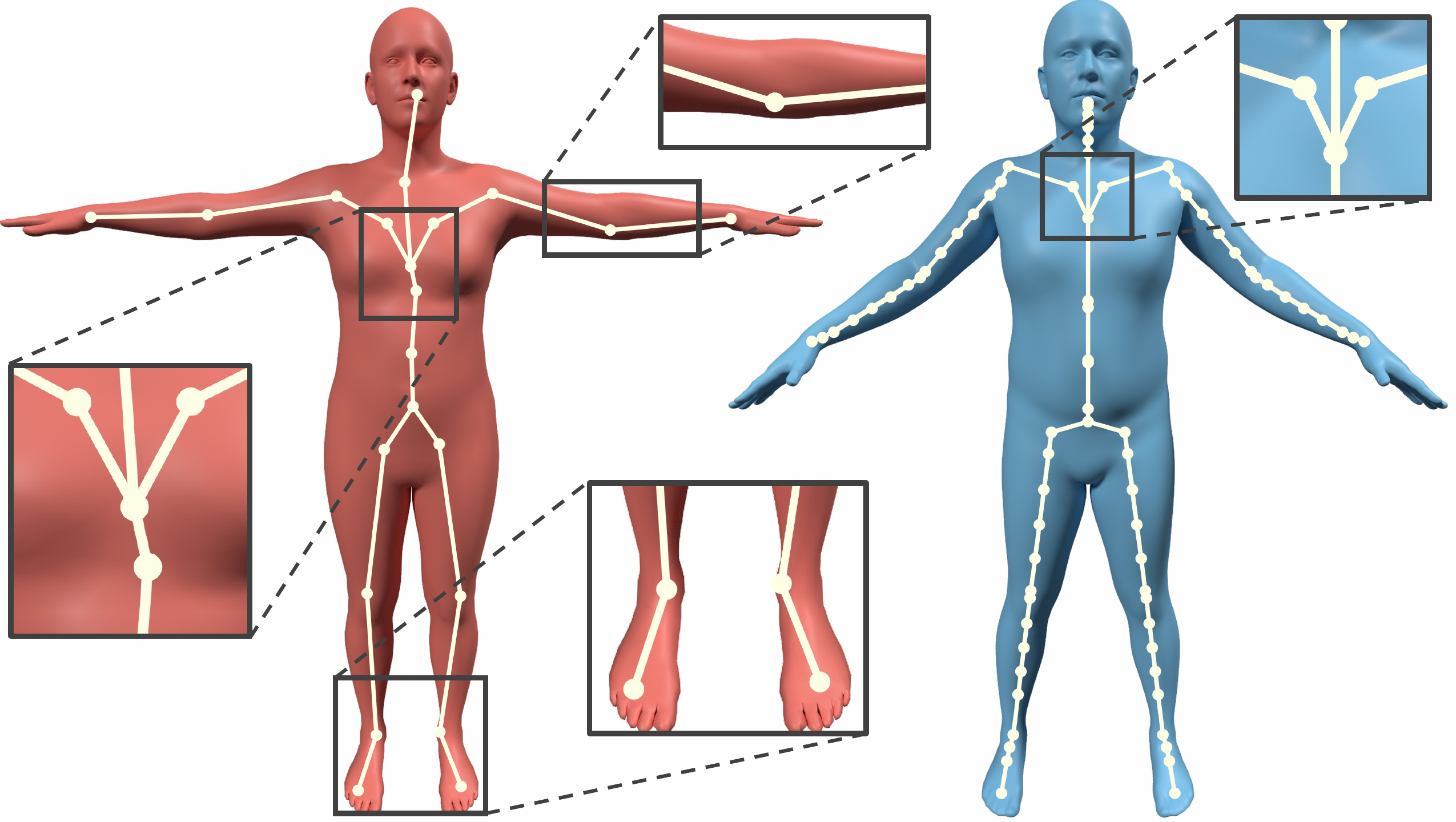}
  \vspace{-0.1in}
  \caption{\textbf{Comparison of Skeleton Symmetries.} (\textit{Left}) The mean SMPL-X mesh reveals significant skeletal asymmetries (elbows, spine, and feet) due to joint centers being derived from vertices. (\textit{Right}) ATLAS mesh demonstrates a symmetric and consistent skeleton through decoupling of skeletal and surface parameters.}

  \label{fig:smplx_joint_movement}
  \vspace{-0.2in}
\end{figure}

Recent years have seen significant advancements in human-centric applications, including 3D digitization of avatars for virtual reality~\cite{lombardi2019neural,saito2020pifuhd,weng2022humannerf,xiu2022icon}, efficient and performant motion capture~\cite{yin2023hi4d,cheng2023dna,peng2021neural,zheng2022structured}, physically plausible human-object interaction~\cite{bhatnagar2022behave,chao2015hico,wang2024review}, and generative human character generation~\cite{peng2024charactergen,ren2023make,liang2024rich}. Supporting these methods are parametric models of the human body~\cite{SMPL:2015,SMPL-X:2019,xu2020ghum,anguelov2005scape,wang2020blsm,yang2020facescape} --- methods that derive diverse, articulated human meshes from low-dimensional shape and pose parameters. Expressive and controllable parametric human models are thus critical for advancing the broad field of human understanding.

The dominant approaches for parametric modeling of the human body follow a vertex-centric framework \cite{SMPL:2015,SMPL-X:2019,STAR:ECCV:2020,osman2022supr,xu2020ghum} where surface vertices are personalized through a linear basis, internal skeletal joints are derived from the surface through weighted sum, and the mesh is driven with linear blend skinning (LBS)~\cite{kavan2007skinning} and pose dependent corrections. While achieving plausible 3D reconstruction, this paradigm presents several inherent limitations. First, deriving internal skeletal joints from surface vertices introduces incorrect correlations. Shown in Figure \ref{fig:smplx_joint_movement}, the skeletal joints in SMPL-X~\cite{SMPL-X:2019} are asymmetrical, and the spine shifts left-to-right with changes in the second shape component which is associated with soft tissue variation. Second, skeletal attributes can only be modified by altering shape components, which inevitably affects other surface vertex attributes. For instance, shoulder width is intertwined with several components in SMPL-X~\cite{SMPL-X:2019} that affect soft tissue, inhibiting precise customization of internal attributes (refer Figure \ref{fig:smplx_shoulder_width}). Third, this correlation causes keypoint fitting to produce meshes with unwarranted soft-tissue deviations, although keypoints provide no information about these attributes.

\newcommand{\easytilde}{\raisebox{0.5ex}{\texttildelow}}

\begin{table}[b]
\small
\begin{center}
\vspace{-0.2in}
\resizebox{\linewidth}{!}{
    \setlength{\tabcolsep}{2pt}
    \setlength\extrarowheight{-1pt}
    \renewcommand{\arraystretch}{1.1}
    \begin{tabular}{lrcccc}
    \toprule
    \textbf{Method} & \textbf{Shape IDs} & \textbf{Pose IDs} & \textbf{Scans}  & \textbf{Shape Basis} & \textbf{Skeletal Basis} \\
    \midrule
    SMPL~\cite{SMPL:2015} & 3.8k & 40 & 1.8k & \cmark & \xmark \\ 
    SMPL-X~\cite{SMPL-X:2019} & 3.8k & 40 & 1.8k & \cmark & \xmark \\ 
    STAR~\cite{STAR:ECCV:2020} & 13k & 40 & 1.8k & \cmark & \xmark \\ 
    BLSM~\cite{wang2020blsm} & 3.8k & 10 & 41k & \cmark & \cmark \\ 
    GHUM~\cite{xu2020ghum} & 4.3k & 48 & 60k & \cmark & \xmark \\ 
    OSSO~\cite{keller2022osso} & 2k & - & 2k & \cmark & \cmark \\ 
    SUPR~\cite{osman2022supr} & 15k & - & 1.2m$^{\dagger}$ & \cmark & \xmark \\ 
    BOSS~\cite{shetty2023boss} & 300 & - & 300 & \cmark & \cmark \\ 
    SKEL~\cite{keller2023skin} & 3.9k & 113 & 1m$^{\dagger}$ & \cmark & \cmark \\ 
    \midrule
    \NAME{} (Ours) & 15k & 157 & 600k & \cmark & \cmark\\ 
    \bottomrule
    \end{tabular}
}
\vspace{-0.1in}
\caption{\textbf{Comparison of training data across state-of-the-art parametric models.} \NAME{} leverages diverse registrations across shape and pose at large scale. $^{\dagger}$SUPR \cite{osman2022supr} and SKEL \cite{keller2023skin} use $60$Hz and $30$Hz scans while \NAME{} uses $5$Hz scans.}
\label{table:introduction}
\vspace{-0.2in}
\end{center}
\end{table}

To address these issues, we propose \textbf{\NAME{}}, an expressive parametric model of the human body which explicitly decouples external shape and internal skeleton. Our model is natively trained at high resolution (115k vertices) and has an anatomically motivated skeleton with $77$ joints. To drive ATLAS, we start with a template, unposed mesh and customize soft-tissue attributes (\eg torso and leg volume \etc) with a linear basis over shape. At this stage, the skeletal joints remain unchanged. We use a skeletal basis to customize the internal skeleton, and then we scale and pose the mesh together with LBS~\cite{kavan2007skinning}. By explicitly decoupling external shape and internal skeleton, \NAME{} eliminates spurious vertex-joint correlations and enables more precise controllability of the human mesh as shown in Figure \ref{fig:smplx_shoulder_width}. Further, to improve the realism of the skinned mesh, we introduce sparse, non-linear pose corrective deformations prior to the LBS phase. The sparsity of this mapping prevents the pose correctives from fitting to spurious correlations in the data, such as actuation of one elbow affecting vertices of the other, and the non-linearity enables more accurate deformations around difficult joints (shoulders, elbow tips, \etc).

\NAME{} is trained on a large-scale dataset of 600k high-resolution scans of minimally clothed subjects in diverse poses. Compared to prior work as shown in Table \ref{table:introduction}, our model is developed from a more diverse set of shapes, identities, and poses, resulting in a more expressive human body model. Additionally, for in-the-wild usage, we develop a single image model fitting pipeline. Our framework leverages \NAME{}'s decoupling of the shape and skeleton, as well as recent advancements in high-fidelity human-centric models~\cite{khirodkar2024sapiens}, to first fit the skeleton and pose to keypoints, then personalize body shape to fit the human silhouette. Supporting the fitting is a VAE pose prior \cite{kingma2013auto,SMPL-X:2019} trained on 600k frames as well as a PCA prior over hand poses. Our evaluations show that the resulting pipeline better fits poses and derives more plausible shape compared to existing methods. To evaluate our model, we provide quantitative results on fitting a diverse dataset of body shapes and poses \cite{3DBodyTex} and demonstrate that \NAME{} is more expressive than existing state-of-the-art parametric body models. 

\noindent
Our contributions are summarized as follows.
\begin{itemize}
\itemsep0em 
  \item We propose \NAME{}, a controllable and expressive parametric human body model with separate bases for external shape and internal skeleton.
  \item Our model, with sparse, non-linear pose correctives, demonstrates more expressivity in representing shaped, articulated human scans.
  \item We leverage our model to build a high-fidelity single RGB image to model parameter pipeline that captures diverse poses, body shapes, and expressions.
\end{itemize}

\section{Related Work}
\noindent\textbf{3D Human Mesh Modeling.}
The early work SCAPE \cite{Anguelov05} separately models pose and shape changes with triangle deformations. Follow-up works refine or constrain the deformations, improve registrations, or apply it to soft-tissue dynamics \cite{hasler2009statistical,hirshberg2012coregistration,freifeld2012lie,chen2013tensor,pons2015dyna}. 
SMPL \cite{SMPL:2015} proposes a vertex-based model \cite{allen2006learning} with shape and pose corrective blendshapes and uses LBS to pose the mesh around joints regressed from vertices. STAR \cite{STAR:ECCV:2020} compacts SMPL by using quaternions and sparsifies the corrective matrix. Frank \cite{joo2018total} adds the FaceWarehouse \cite{cao2014facewarehouse} face model and an artist-designed hand model to the SMPL rig. SMPL-H \cite{romero2017embodied} adds MANO, a hand model learned from scans, and SMPL-X \cite{SMPL-X:2019} merges MANO, FLAME, and SMPL to learn a shape space for the entire body. SUPR \cite{osman2022supr} improves on SMPL-X with using a federated dataset of body parts and a better foot model. GHUM \cite{xu2020ghum} proposes a non-linear shape space and pose correctives. Following the SMPL framework of regressing joints from surface vertices, these works introduce suboptimal correlations between external shape and internal skeleton. In contrast, ATLAS explicitly learns a separate skeleton space for better controllability and shape expressivity.

\begin{figure}[t]
  \centering
  \includegraphics[width=\linewidth]{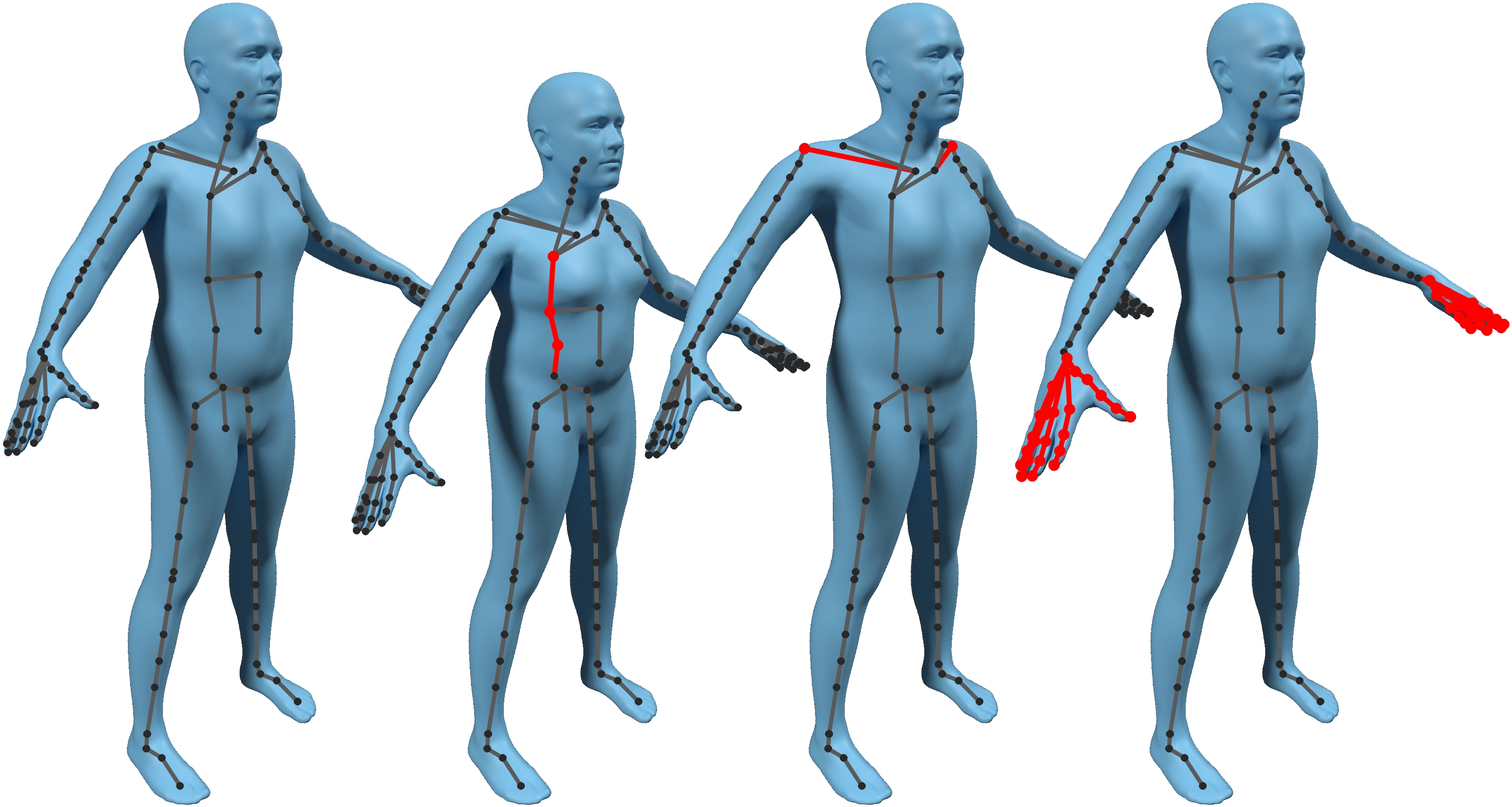}
  \vspace{-0.2in}
  \caption{\textbf{Controllable Skeletal Attributes.} From left to right, we visualize the template mesh, decreasing spine length, increasing shoulder width, and increasing scale of both hands.}
  \label{fig:scale_demo}
  \vspace{-0.2in}
\end{figure}

\vspace{1mm}
\noindent\textbf{Skeleton Models.}
Many works in biomechanics define precise, anatomically accurate skeleton models. Some works \cite{rajagopal2016full,seth2016biomechanical,nitschke2020efficient} derive musculoskeletal models for modeling anatomically plausible movement. Other methods \cite{Dicko2013AnatomyT,gilles2010creating,kadlevcek2016reconstructing,saito2015computational,zhu2015adaptable} optimize for underlying fat, muscle, and bone that drive surface deformations. While anatomically constrained, these methods rely on specialized simulators \cite{lee2009comprehensive} or use models for bone or fat growth, making their usage in commercial graphics packages difficult. OSSO \cite{keller2022osso} and BOSS \cite{shetty2023boss} derive the anatomical skeleton from SMPL meshes from medical segmentation masks, but driving them to new poses requires additional optimization.

Most related to our work are methods that introduce decoupled skeleton-driven human mesh models for graphics and commercial applications. An early work \cite{hasler2010learning} optimizes internal bones for both pose changes and shape variation. In contrast, we maintain a separate space over vertices to represent shape, better modeling soft tissue deformations. BLSM \cite{wang2020blsm} proposes a decoupled shape and bone-length space. While promising, the model is not open and lacks pose corrective deformations which limits realism. SKEL \cite{keller2023skin} places bony and soft markers on AMASS \cite{AMASS:ICCV:2019} using OSSO \cite{keller2022osso} and uses AddBiomechanics \cite{werling2022rapid} to optimize for an internal skeleton. SKEL then learns a mapping from vertices joints and re-rigs the SMPL mesh. While effective in deriving biomechanical skeletons from SMPL meshes, SKEL inherits SMPL's surface vertex-based shape space to synthesize new human meshes. Skin generation for a specified skeleton requires optimization to find matching SMPL shape parameters, and it requires that the desired skeleton is represented in the SMPL shape space. Further, SKEL lacks finger control and inherits a limited set of pose correctives from SMPL. In contrast, our ATLAS enables direct controllability of decoupled shape and scale parameters, includes finger actuation, and provides expressive pose correctives.

\vspace{1mm}
\noindent\textbf{Pose Corrective Deformations.}
To model pose-dependent deformations, Lewis \cite{Lewis:2000:PSD} applies vertex offsets around joints to alleviate ``collapsing joint'' defects. Other works \cite{Allen:TOG:2002,kurihara2004modeling,rhee2006real} interpolate between saved deformations for key poses and EigenSkin \cite{kry2002eigenskin} adds a PCA space for each joint. SMPL \cite{SMPL:2015} learns a mapping from joint rotations to vertex deformations, and STAR \cite{STAR:ECCV:2020} sparsifies this mapping through geodesic initialization and regularization. GHUM \cite{xu2020ghum} improves the expressiveness of this mapping through a non-linear network. However, by mapping the full body pose to a compressed 32-dim intermediate latent vector, GHUM's pose correctives remain dense. Our ATLAS seeks to achieve the best of both worlds through a sparse and non-linear mapping, avoiding spurious correlations while maintaining the expressiveness of non-linear correctives.

\section{Method}

In this section, we present the ATLAS model, detail its controllable skeletal attributes, describe our sparse non-linear pose-correctives, and present a single-image fitting method.

\subsection{Decoupled Skeletal and Shape Body Model}\label{method:model}
\jp{\noindent \textbf{Overview.} The core strength of ATLAS lies in its explicit separation of the external surface from the internal skeleton. This characteristic enables precise, independent customization of surface and skeletal attributes. To support this capability, ATLAS derives a shaped and posed human mesh in two steps. \textbf{First}, surface vertices are customized while aligned to a fixed template skeleton. \textbf{Second}, this mesh is simultaneously scaled and posed using LBS \cite{kavan2007skinning}, modifying the underlying skeleton using 76 individually controllable skeletal attributes that each control bone lengths and body part sizes (Section \ref{method:skeleton}). The resulting mesh accurately captures subtle variations of the human shape.}

\begin{figure}[t]
  \centering
  \includegraphics[width=\linewidth]{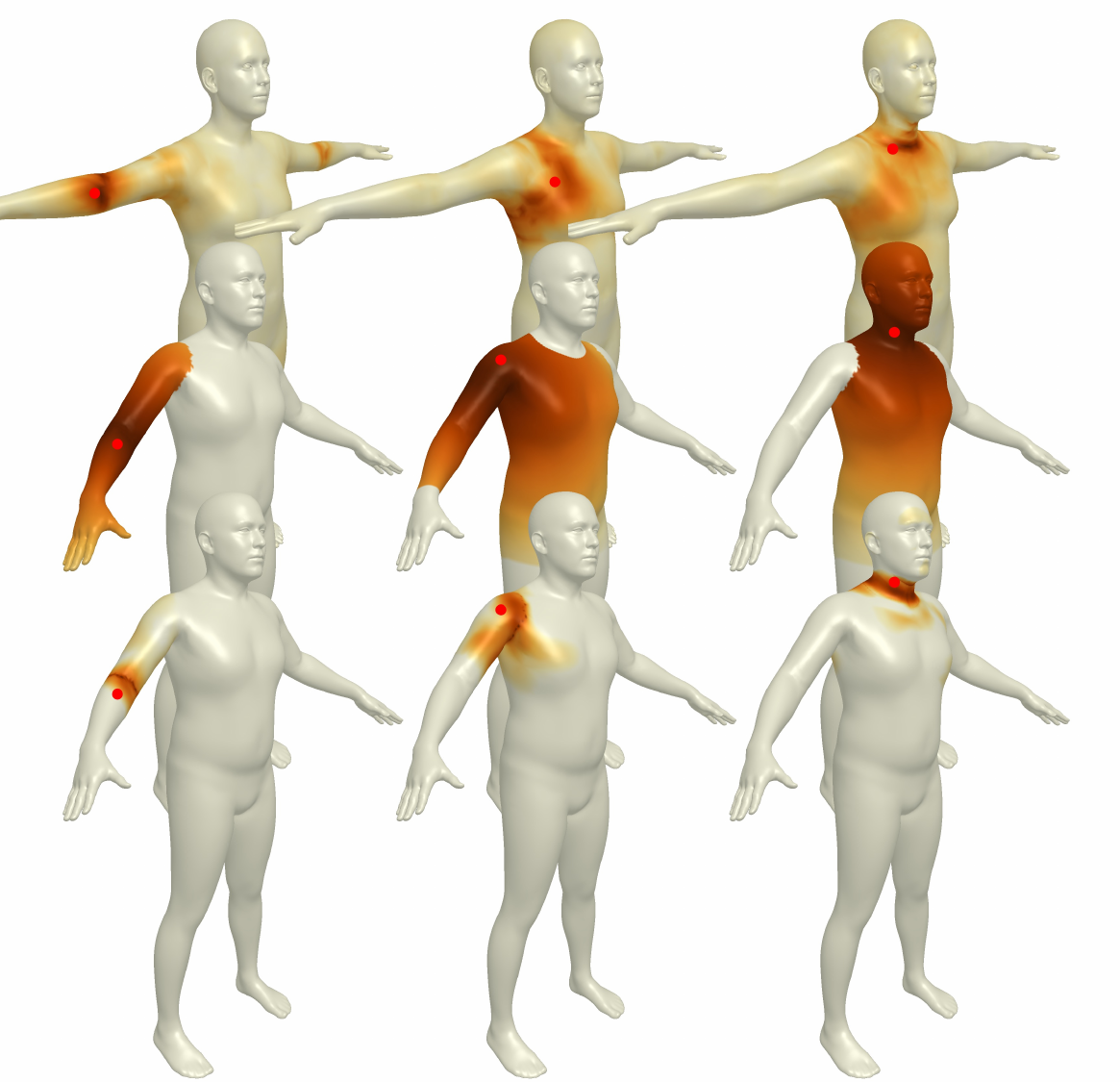}
  \vspace{-0.2in}
  \caption{\textbf{Sparse Pose Correctives.} The first row displays pose correctives from SMPL-X. The second row shows the inverse geodesic initialization for our pose corrective activations, and the third row demonstrates their sparsity after convergence.}
  \label{fig:pcb_trained}
  \vspace{-0.2in}
\end{figure}

\noindent \jp{\textbf{Surface Customization.}
To shape and pose an ATLAS mesh, we first obtain customized surface vertices $\tilde{X}$ aligned to the template skeleton in the default A-pose:}
\begin{equation}
\tilde{X}(\beta^s, \beta^f, \theta) = \bar{X} + \mathcal{B}^s(\beta^s, \mathcal{S}) + \mathcal{B}^f(\beta^f, \mathcal{F}) + \mathcal{B}^p(\theta, \mathcal{P})
\end{equation}
where $\bar{X} \in \mathbb{R}^{3V}$ is the template A-pose shape, $\mathcal{B}^s(\beta^s, \mathcal{S}) = \sum_{n=1}^{|\beta^s|} \beta^s_n\mathcal{S}_n$ is the surface vertices' blend shape function, and $\mathcal{B}^f(\beta^f, \mathcal{F}) = \sum_{n=1}^{|\beta^f|} \beta^f_n\mathcal{F}_n$ is the facial expressions blend shape function. To correct artifacts caused by LBS~\cite{kavan2007skinning}, we add pose deformations $\mathcal{B}^p(\theta, \mathcal{P})$. Unlike prior works~\cite{SMPL:2015,SMPL-X:2019} that derive joint centers from this customized identity shape, our mesh at this stage remains unposed, \textit{unscaled}, and aligned to a fixed internal skeleton. 

\vspace{1mm}
\noindent \jp{\textbf{Skeleton Customization.}}
\jp{Next, $\tilde{X}$ is both posed and scaled through LBS~\cite{kavan2007skinning}. During this, the skeleton is customized using $N_k = 76$ controllable attributes: 15 modify body part sizes and 61 adjust bone lengths (Sec. \ref{method:skeleton} and Fig. \ref{fig:scale_demo}).} We denote these as $\ell = \sigma \oplus t$ where $\ell \in \mathbb{R}^{N_k}$ consists of scale $\sigma \in\mathbb{R}^{15}$ and $t \in \mathbb{R}^{61}$ bone length modifications. While these attributes can be set individually, we also learn a blend shape function over them $\mathcal{B}^k(\beta^k) = \sum_{n=1}^{|\beta^k|} \beta^k_n\mathcal{K}_n$ with $\mathcal{K}_n \in \mathbb{R}^{N_k}$ to capture common variations. 

ATLAS is driven by Euler 3DoF poses $\theta \in \mathcal{R}^{3(J + 1)}$ and skin weights $\omega \in \mathbb{R}^{V\times I}$, where each vertex is affected by up to $I = 8$ joints. Altogether, the surface vertices $\tilde{X}$ are then driven by the modified skeleton and the pose using the LBS function $M$:
\begin{equation}
X(\beta, \theta) = M(\tilde{X}(\beta^s, \beta^f, \theta), \mathcal{B}^k(\beta^k), \theta, \omega) 
\end{equation}
We emphasize that in ATLAS, the joint locations used for posing are \textit{independent} of the vertex components $\beta^s$, which are only used for specifying the A-pose shape $\tilde{X}$. Rather, only the skeletal components $\beta^k$ and the pose $\theta$ specify the joint locations, enabling precise decoupling of the surface and the skeleton. We refer to Section \ref{sec:skinning} of the supplement for a detailed mathematical formulation of $M$.

\subsection{Controllable Skeletal Attributes}\label{method:skeleton}
\jp{ATLAS incorporates $N_k = 76$ skeletal attributes, including 15 scale modifications that directly alter the overall size of body parts (body size, head, hands, feet, and individual fingers)} and 61 bone length parameters that adjust joint translations relative to their kinematic parents. These bone length attributes encompass major body bones including spine, neck, upper and lower arms, upper and lower legs, and fingers. \jp{Figure \ref{fig:scale_demo} demonstrates some examples of individual scale attribute control, and visualizations of their effects are in Figure \ref{fig:skeleton_details} of the supplementary.}

\subsection{Sparse, Non-Linear Pose Correctives}\label{method:pcb}
\jp{\noindent \textbf{Overview.} Pose-dependent deformations before LBS~\cite{kavan2007skinning} are crucial for realistic meshes.
Prior work highlight the benefits of both sparse-linear correctives~\cite{STAR:ECCV:2020,osman2022supr} -- which restrict joint influences to local vertices to avoid spurious correlations -- and dense non-linear correctives~\cite{xu2020ghum}, where each joint non-linearly contributes to all vertices. We reconcile these methods by introducing sparse non-linear correctives that leverage the locality of sparse-linear approaches while preserving the expressivity of non-linear correctives.}

\vspace{1mm}
\noindent \textbf{Pose Correctives Formulation.} Our correctives function $\mathcal{B}^p(\theta, \mathcal{P}) \in \mathbb{R}^{6J} \rightarrow{} \mathbb{R}^{3V}$ takes joint angles in 6D~\cite{zhou2019continuity} and outputs vertex offsets. \jp{We decompose $\mathcal{B}^p$ into a local, non-linear operation and a sparse, geodesic-initialized linear operation. The former encodes local joint groups together, effectively enabling non-linear expressivity, while the latter constrains the extent of vertices each joint group can affect, avoiding spurious joint-vertex correlations.}

\jp{First, the local, non-linear operation processes pose angles of joint $j$ and those of its immediate immediate kinematic neighbors $n(j)$ using a lightweight MLP:
\begin{small}
\begin{equation}
    \text{Non-Linear}_j(\theta) = \text{MLP}\left( \{R_{6d}(\theta_a) - R_{6d}(\vec{0}) \mid a \in n(j)\} \right),
\end{equation}
\end{small}
yielding a $c$-dimensional feature that encodes their poses.}

\jp{This local feature is then transformed into pose corrective vertex offsets around $j$ using a learned mapping:}
\begin{equation}
    \mathcal{B}^p_j = \phi(A_j) \odot \left(P_j \times \text{Non-Linear}_j(\theta)\right)
\end{equation}
\jp{Here, \(P_j \in \mathbb{R}^{3V \times c}\) is the pose corrective weight matrix, and the multiplication \(P_j \times \text{Non-Linear}_j(\theta)\) produces the raw pose-dependent vertex offsets. Following STAR~\cite{STAR:ECCV:2020}, the function \(\phi\) is a ReLU applied to the joint mask \(A_j \in \mathbb{R}^V\) to enforce vertex sparsity per joint.}
For vertex $i$, we initialize the $i$-th element of $A_j$ as $(1 - d(i, j)) \boldsymbol{1}_{i \in \text{seg}(j)}$, where $d(i, j)$ is the normalized geodesic distance from vertex $i$ to the vertex ring around $j$, and $\boldsymbol{1}_{i \in \text{seg}(j)}$ indicates if vertex $i$ belongs to joint $j$'s corresponding or adjacent body part. This initialization, coupled with L1 regularization on $\phi(A)$, encourages sparsity in activation. Figure \ref{fig:pcb_trained} shows the activation mask pre- and post-training, showing pose correctives concentrated around the actuated joint. \jp{Altogether, our pose correctives integrate the expressivity of non-linear functions with the sparsity of regularized linear correctives.}

\subsection{Single-Image Mesh Fitting Application}\label{method:fitting}
\jp{We demonstrate the applicability of ATLAS to real-world images by developing a single-image mesh fitting pipeline.}

\vspace{1mm}
\noindent\textbf{Objective.} Our framework improves upon previous approaches~\cite{Bogo:ECCV:2016,SMPL-X:2019} by explicitly decoupling skeleton and shape fitting while leveraging predictions from high-fidelity human-centric models~\cite{khirodkar2024sapiens}. We optimize shape, skeleton, and pose parameters using the objective~\cite{Bogo:ECCV:2016,SMPL-X:2019}:
\begin{small}
\begin{equation}
\vspace{-.5em}
    E(\beta^s, \beta^f, \beta^k, \theta) =~ E_{data} + E_{\theta_{body}} + E_{\theta_{hand}} + E_{\beta^s} + E_{\beta^f} + E_{\beta^k}
\end{equation}
\end{small}

$E_{data}$ comprises of three components: $E_{kps2d} + E_{depth} + E_{mask}$. $E_{kps2d}$ minimizes the distance between projected 3D ATLAS keypoints and 2D detector keypoints using a robust loss~\cite{geman1987statistical}. $E_{depth}$ minimizes differences between rendered mesh depth and Sapiens~\cite{khirodkar2024sapiens} relative depth predictions. $E_{mask}$ uses efficient Edge Gradients~\cite{Pidhorskyi2024RasterizedEG} to minimize differences between rendered and predicted foreground masks. For pose regularization, we train a VAE prior on full-body poses (excluding hands) and optimize in the latent space with L2 prior $E_{\theta_{body}}$. Hand poses are optimized in PCA 6D ~\cite{zhou2019continuity} space with L2 prior $E_{\theta_{hand}}$. Similarly, we apply L2 regularization to shape, expression, and skeleton attribute latents via $E_{\beta^s}$, $E_{\beta^f}$, and $E_{\beta^k}$. 

\begin{figure}[b]
  \centering
  \vspace{-0.2in}
  \includegraphics[width=\linewidth]{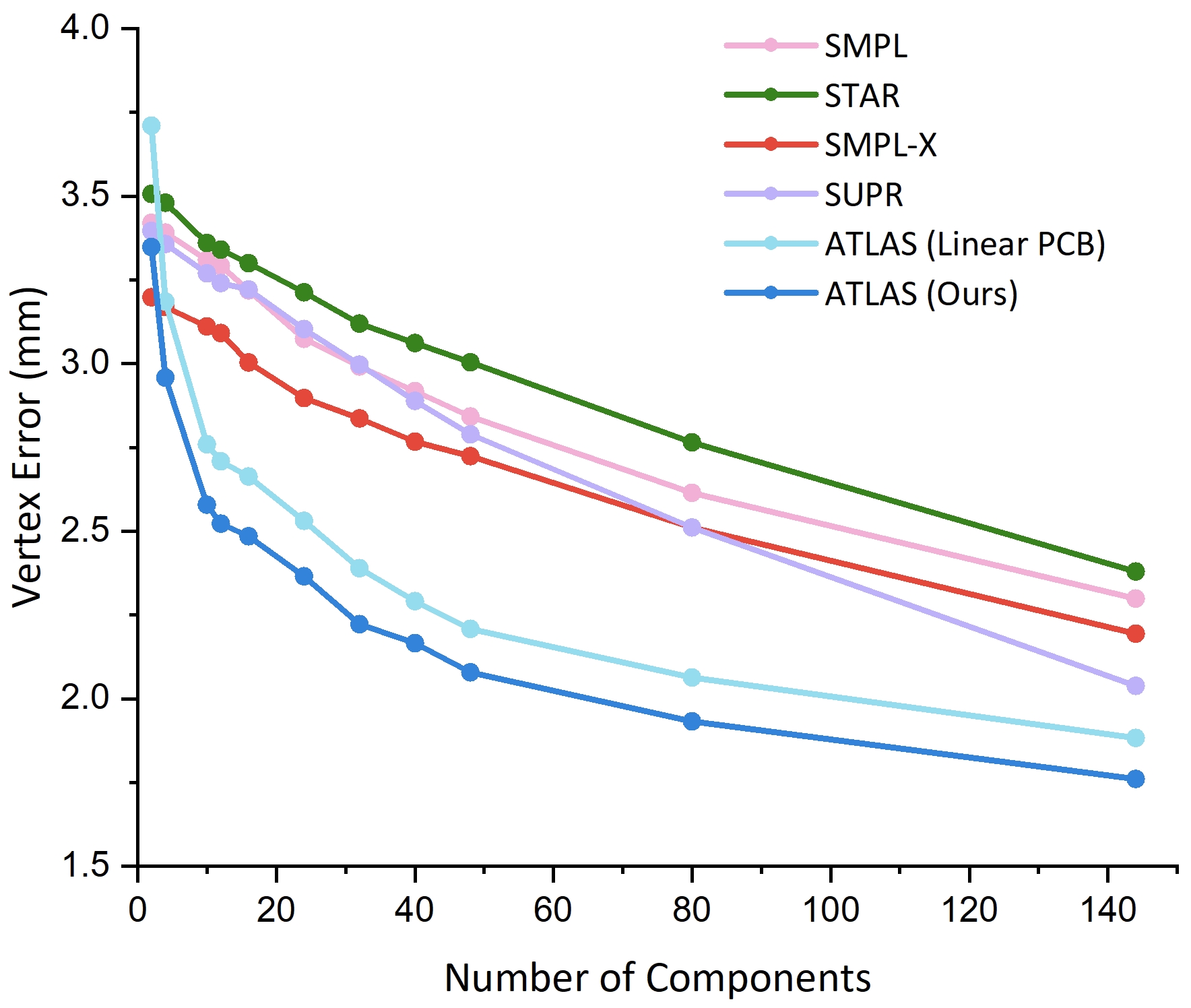}
  \vspace{-0.2in}
  \caption{\textbf{Quantitative Evaluation on 3DBodyTex.} We report vertex-to-vertex error (mm) with different numbers of fitting components. For ATLAS, we report the combination of the number of shape and scale components used.}
  \label{fig:3dbodytex_quant}
  \vspace{-0.2in}
\end{figure}

\vspace{1mm}
\noindent\textbf{Optimization.} We use multi-stage optimization, first fitting major body keypoints followed by hands and expressions. We explicitly decouple skeleton and shape parameter optimization: skeleton latents $\beta^k$ are optimized using only pose and skeletal structure of the subject, $E_{kps2d} + E_{depth}$, while surface shape latents $\beta^s$ are optimized using the mask term $E_{mask}$. This separation enables clean optimization of pose and body structure through keypoint and depth terms, while accurately capturing soft tissue variations through mask fitting. Unlike prior work~\cite{SMPL-X:2019} that entangles skeleton and shape, our approach prevents keypoint-induced soft tissue hallucinations while fitting subject silhouettes through shape variation. For facial expression fitting, we introduce an improved approach that first aligns projected 3D expression keypoints with target 2D keypoints using optimal rotation and translation, then minimizes their difference using a robust loss term. This enables realistic expression capture even with head misalignment.

\begin{figure*}[t]
  \centering
  \includegraphics[width=\linewidth]{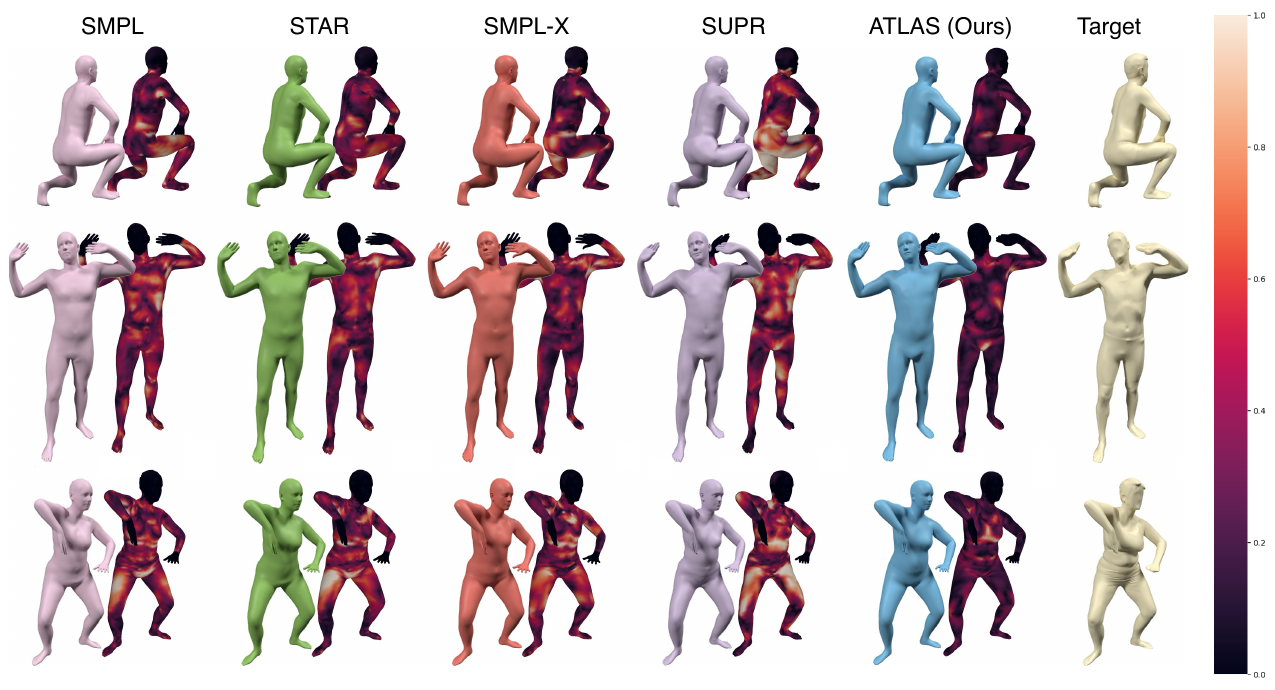}
  \vspace{-0.4in}
  \caption{\textbf{Qualitative Results on 3DBodyTex.} We visualize each model fit with $16$ components, as well as a heatmap indicating vertex-to-vertex error. Overall ATLAS exhibits tighter fits and has fewer blending artifacts at the elbows, knees, and shoulders compared to baselines.}
  \label{fig:3dbodytex}
  \vspace{-0.2in}
\end{figure*}

\section{Experiments}
In this section, we first describe ATLAS training, followed by an extensive comparison with state-of-the-art body models across multiple scenarios. Lastly, we provide insights into the importance of pose correctives in ATLAS and demonstrate its fitting to in-the-wild images.

\begin{figure}[b]
  \vspace{-0.2in}
  \centering
  \includegraphics[width=\linewidth]{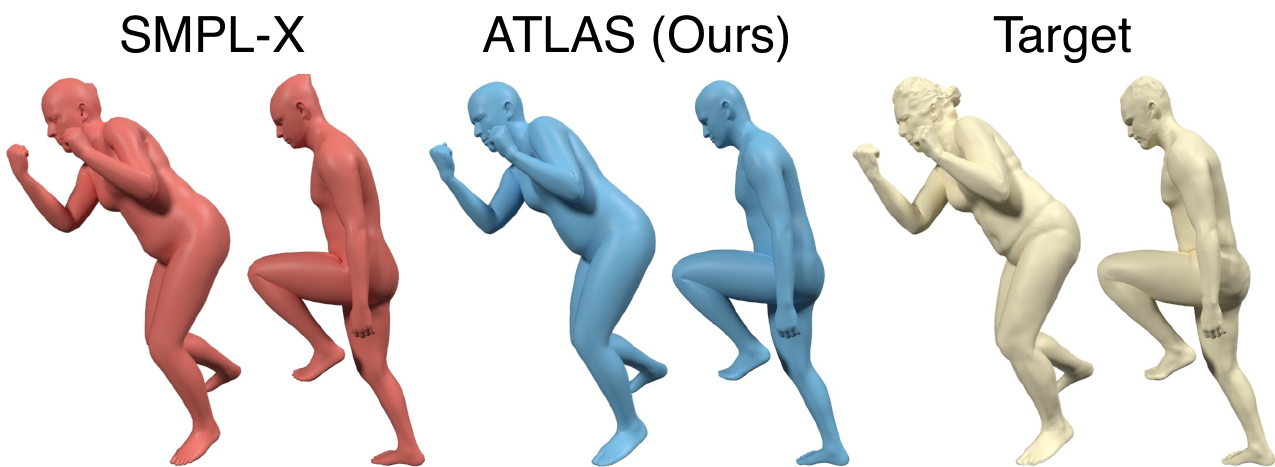}
  \vspace{-0.22in}
  \caption{\textbf{Qualitative Results on the Goliath-Test set.} From the left, we compare SMPL-X's fits, ATLAS's fits, and registrations. ATLAS is noticeably better at capturing areas around joints, resulting in sharper knees and elbows.}
  \label{fig:internal_qual}
  \vspace{-0.22in}
\end{figure}

\begin{figure}[b]
  \centering
  \vspace{-0.2in}
  \includegraphics[width=\linewidth]{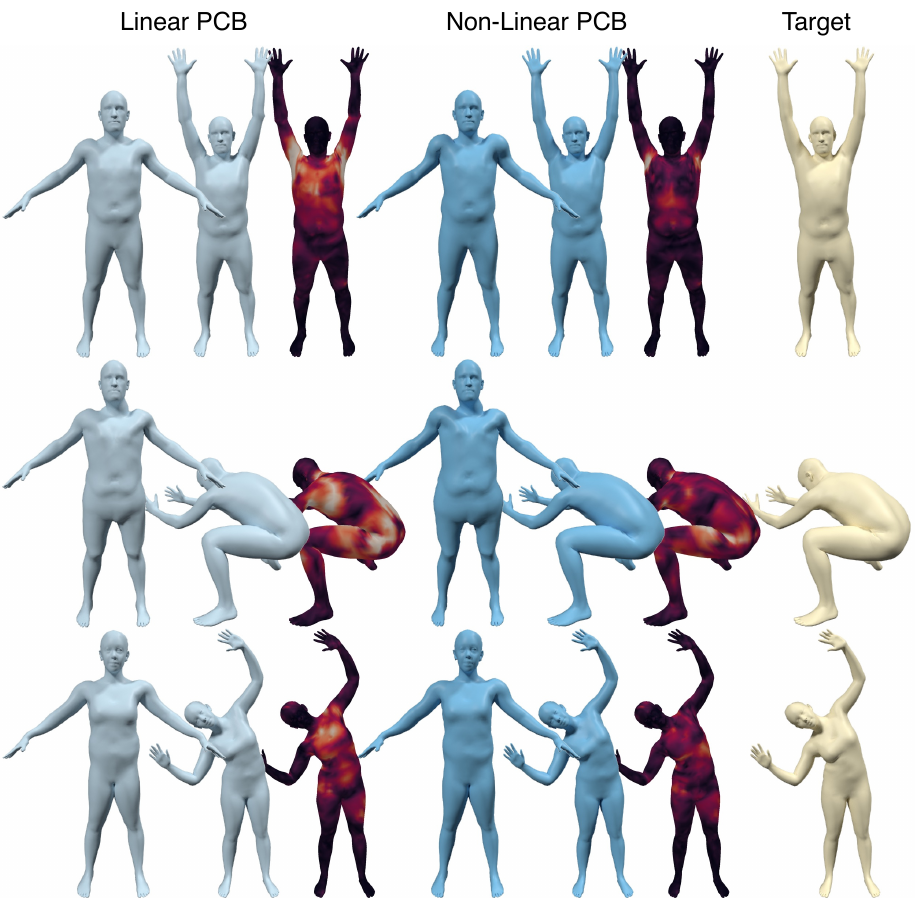}
  \vspace{-0.2in}
  \caption{\textbf{Qualitative comparison of linear and non-linear pose correctives (PCB) on the SMPL pose dataset.} For each of linear and non-linear PCBs, we visualize the predicted vertex offsets in the rest pose, the posed mesh, and the fitting error heatmap.}
  \label{fig:pcb_eval}
  \vspace{-0.1in}
\end{figure}

\subsection{Training ATLAS}\label{exp:datamodel}
\noindent
\textbf{Goliath Dataset.} We collect a dataset of $600K$ high-resolution scans from $130$ subjects in dynamic poses, named \textit{Goliath} for its scale. These scans are captured with a calibrated and synchronized multi-view camera system with $240$ cameras at 4K resolution. Figure~\ref{fig:internal_qual} shows images of participating subjects in our capture setup. Notably, this dataset is substantially larger than existing datasets; \eg SMPL~\cite{SMPL:2015} consists of $1.2K$ scans from $27$ subjects. Our dataset's scale enables learning a more generalizable human body model. Additionally, following~\cite{SMPL-X:2019}, we also use existing datasets to train ATLAS, including CAESAR~\cite{CAESAR} and SizeUSA~\cite{SizeUSA} processed by Meshcapade, consisting of $4391$ and $10123$ scans respectively. This dataset captures diverse body shapes, represent a broader section of the population (aged 18 to 65+), and complement our captured data.

\begin{figure*}[h]
  \centering
  \includegraphics[width=\linewidth]{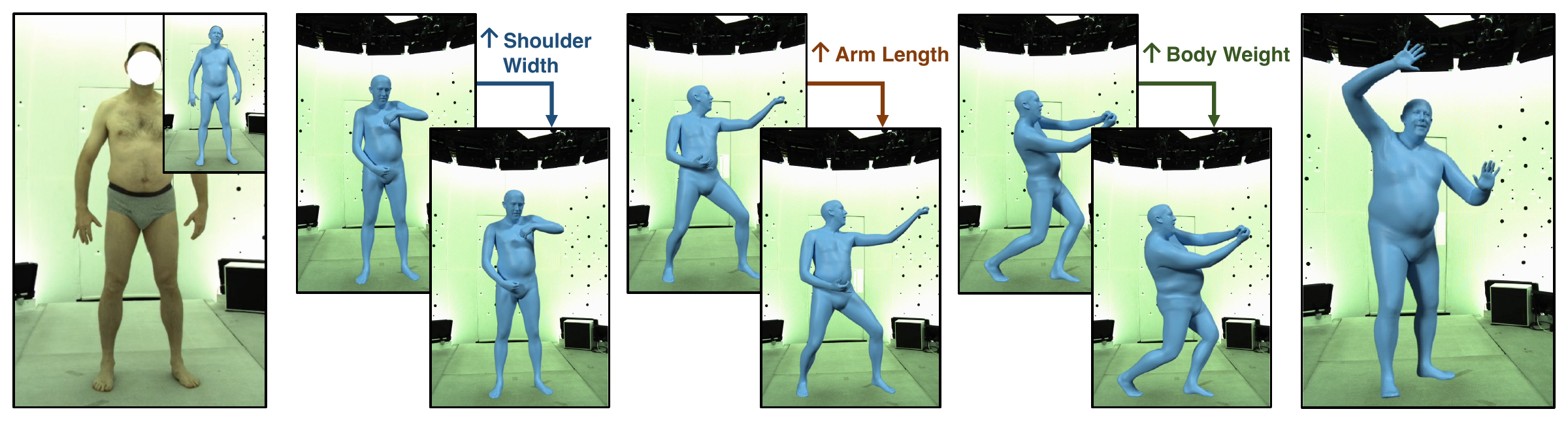}
  \vspace{-0.2in}
  \caption{\jp{\textbf{Precise, decoupled control of skeletal and surface attributes.} Starting with a detailed ATLAS mesh of a subject, we sequentially increase shoulder width, change arm length, and adjust body weight. The resulting mesh is highly realistic, and it maintains details of the original subject shape and skeleton while naturally incorporating the added customizations.}}
  \label{fig:control_demo}
  \vspace{-0.2in}
\end{figure*}

\vspace{1mm}\noindent
\textbf{Body Model.} We design ATLAS to support multiple mesh resolutions. At the highest resolution, the ATLAS mesh consists of 115,834 vertices, approximately 16 times more than the 6,890 vertices in an SMPL~\cite{SMPL:2015} mesh. Our lowest resolution defaults to the SMPL resolution. We train $128$ and $16$ components for the surface vertex space and the skeletal space, respectively and our pose-corrective features are $24$ dimensional. Additionally, we extract a hand pose PCA space, and re-target expressions from FLAME~\cite{FLAME:SiggraphAsia2017}. 

\noindent\textbf{Implementation Details.} We decouple the surface and the skeleton in our data by first optimizing registrations with only skeletal parameters and poses. Using triangulated keypoints to regularize joints, these skeletal-only fits capture variations in height, arm length, finger size, etc. Then, we optimize surface shape to model soft tissue attributes like body weight and arm width. We then train ATLAS to capture these skeletal and surface spaces with autoencoders. Please refer to Section \ref{sec:training_details} of the supplement for more details.

\subsection{Comparison with existing Body Models}
We compare ATLAS with state-of-the-art body models, including SMPL~\cite{SMPL:2015}, STAR~\cite{STAR:ECCV:2020}, SMPL-X~\cite{SMPL-X:2019} and SUPR~\cite{osman2022supr} on two datasets: 3DBodyTex~\cite{3DBodyTex} and Goliath-Test, a held-out test set of our captured dataset. In contrast to these baselines, only our proposed ATLAS body model decouples the skeletal and shape spaces.

\vspace{1mm}\noindent
\textbf{3DBodyTex}~\cite{3DBodyTex}. The dataset consists of $100$ male and $100$ female scans. For evaluation, we register each body model to the ground-truth 3D scans using the SMPL~\cite{SMPL:2015} topology to ensure a fair comparison. Due to missing or noisy data in the ground-truth face and hand regions, we mask these areas during evaluation.

\begin{table}[b]
\begin{center}
\resizebox{3in}{!}{
    \renewcommand{\arraystretch}{1.2}
    \setlength{\tabcolsep}{3pt}
    \begin{tabular}{lrc}
    \toprule
    \textbf{Method} & \textbf{\# Vertices} & \textbf{Runtime (ms)} \\
    \midrule
    SMPL-X & 10475 & 3.74 \\
    ATLAS (SMPL topology) & 6890 & 2.39 \\
    ATLAS (SMPL-X topology) & 10475 & 2.47 \\
    ATLAS (High-resolution) & 115834 & 5.37 \\
    \bottomrule
    \end{tabular}
}
\vspace{-0.1in}
\caption{Runtime comparison of mesh skinning on an A100 GPU.}
\label{table:runtime}
\end{center}
\end{table}

Figure~\ref{fig:3dbodytex_quant} shows the vertex error of all body models with respect to the number of fitting components. ATLAS achieves lower fitting error with fewer components due to its explicit decoupling of skeletal and shape spaces. \jp{For instance, at 32 components, ATLAS achieves 21.6\% lower vertex-to-vertex error compared to SMPL-X. This validates ATLAS's ability to generalize to unseen identities.} Figure~\ref{fig:3dbodytex} shows a qualitative comparison of ATLAS with existing baselines on the 3DBodyTex dataset. We observe that ATLAS especially performs well at the tip of the actuated joints (elbows and knees) and fits the shoulders of the target scan more closely compared to SMPL-X~\cite{SMPL-X:2019}.

\vspace{1mm}\noindent
\textbf{Goliath-Test}. We evaluate on $100$ unseen 3D scans in unique poses from $10$ held-out subjects. Our evaluation protocol remains similar to 3DBodyTex but include the face and hands. The qualitative results are shown in Figure \ref{fig:internal_qual}. \jp{In addition to having sharper joints, ATLAS better captures subtle deformations of the clenched hand and angled chin, and it achieves a lower fitting error of $2.34$ mm compared to SMPL-X's $2.78$ mm.}

\subsection{Discussion}
\jp{\textbf{Controllability.} Compared to prior work, ATLAS's separation of surface and skeleton allows for precise control over the human mesh. In Figure \ref{fig:control_demo}, we demonstrate customization of the ATLAS mesh of a subject. We easily control shoulder width and arm length by adjusting a single skeletal attribute each, then adjust body weight by updating the first surface component. Changes in skeletal attributes precisely maintain the original surface details, and changes in surface attributes keep the internal skeleton constant. The resulting mesh is realistic and can readily be driven by the subject's own motion or by pose sequences from other sources. We encourage readers to refer to the supplemental video.}

\begin{table}[b]
\small
\setlength{\tabcolsep}{2pt}
\renewcommand{\arraystretch}{1.2}
\begin{tabularx}{\linewidth}{l|cc}
\toprule
\textbf{Method} & \textbf{Vertex Error} (mm) & \textbf{Joint Error} (mm) \\
\midrule
SMPLify-X & 87.7 & 73.2 \\
\textbf{ATLAS (Ours)} & \textbf{55.4} & \textbf{53.7} \\
~~~~no rel. depth & 60.7 & 54.5 \\
~~~~no rel. depth, no mask & 61.8 & 55.7 \\
\bottomrule
\end{tabularx}
  \vspace{-0.1in}
\caption{\textbf{Evaluating mesh prediction from a single image.} Our model better predicts the 3D human mesh from a single image, and each data term improves fitting.}\label{tab:2dkps}
  \vspace{-0.1in}
\end{table}

\vspace{1mm}\noindent
\textbf{Linear vs Non-Linear Pose Correctives.} Unlike existing methods, ATLAS uses non-linear pose correctives, which introduces more parameters but provides higher capacity to model pose-correlated vertex corrections. To evaluate, we compare ATLAS against a version with linear pose correctives on the SMPL~\cite{SMPL:2015} dataset, isolating the influence of pose-corrective blendshapes by fitting a single rest-pose mesh and internal skeleton across the entire sequence. A qualitative comparison is shown in Figure~\ref{fig:pcb_eval}. The non-linear correctives achieve more realistic fitting, particularly around complex joints such as the shoulders, and better capture muscle bulging in extreme poses. Quantitatively, the fitting error decreases from $1.82$ mm to $1.61$ mm, with improvements concentrated around joint locations.

\vspace{1mm}\noindent
\textbf{Computational Analysis.} We compare the cost of generating a 3D mesh from the body model parameters in Table~\ref{table:runtime}. ATLAS achieves significantly faster inference times than SMPL-X~\cite{SMPL-X:2019} for the same number of vertices, leveraging its optimized CUDA-based implementation. Moreover, our model supports higher resolutions (10$\times$ more vertices) with minimal latency increase.

\begin{figure*}[t]
  \centering
   \includegraphics[width=1\linewidth]{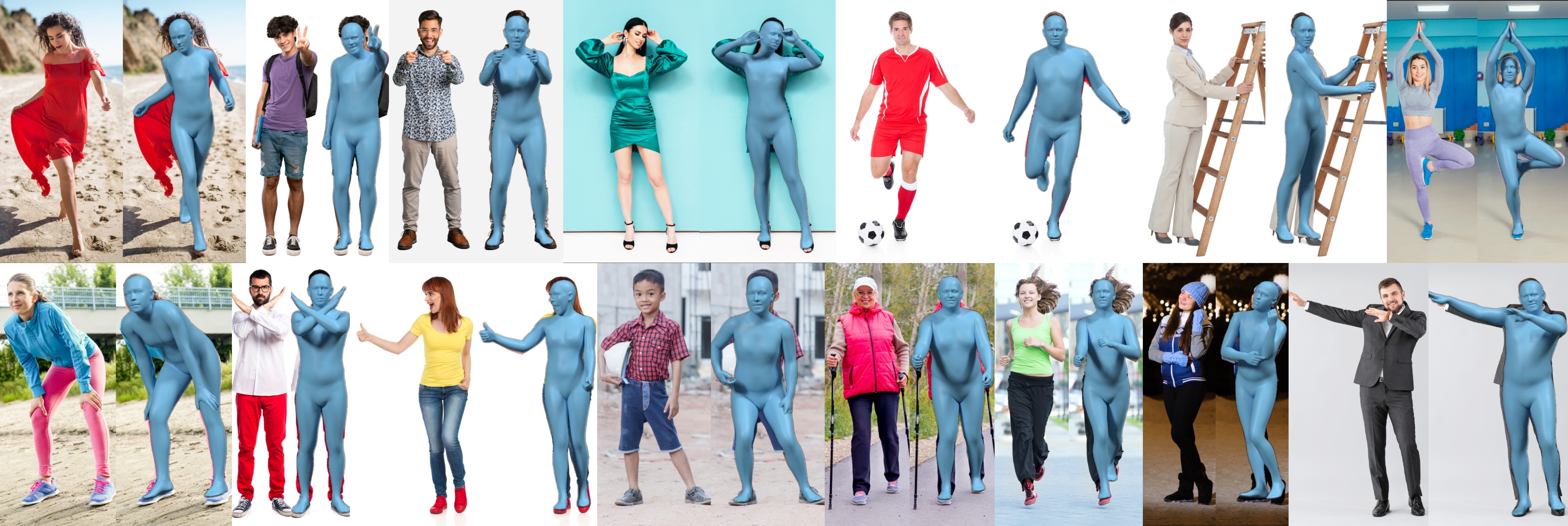}
  \vspace{-0.2in}
  \caption{\textbf{Qualitative results of fitting ATLAS to in-the-wild images.} Our multi-stage body fitting procedure robustly handles clothed subjects in varying poses along with detailed facial expressions.}
  \label{fig:shutterstock}
  \vspace{-0.3in}
\end{figure*}

\begin{figure}[b]
  \centering
  \vspace{-0.2in}
  \includegraphics[width=\linewidth]{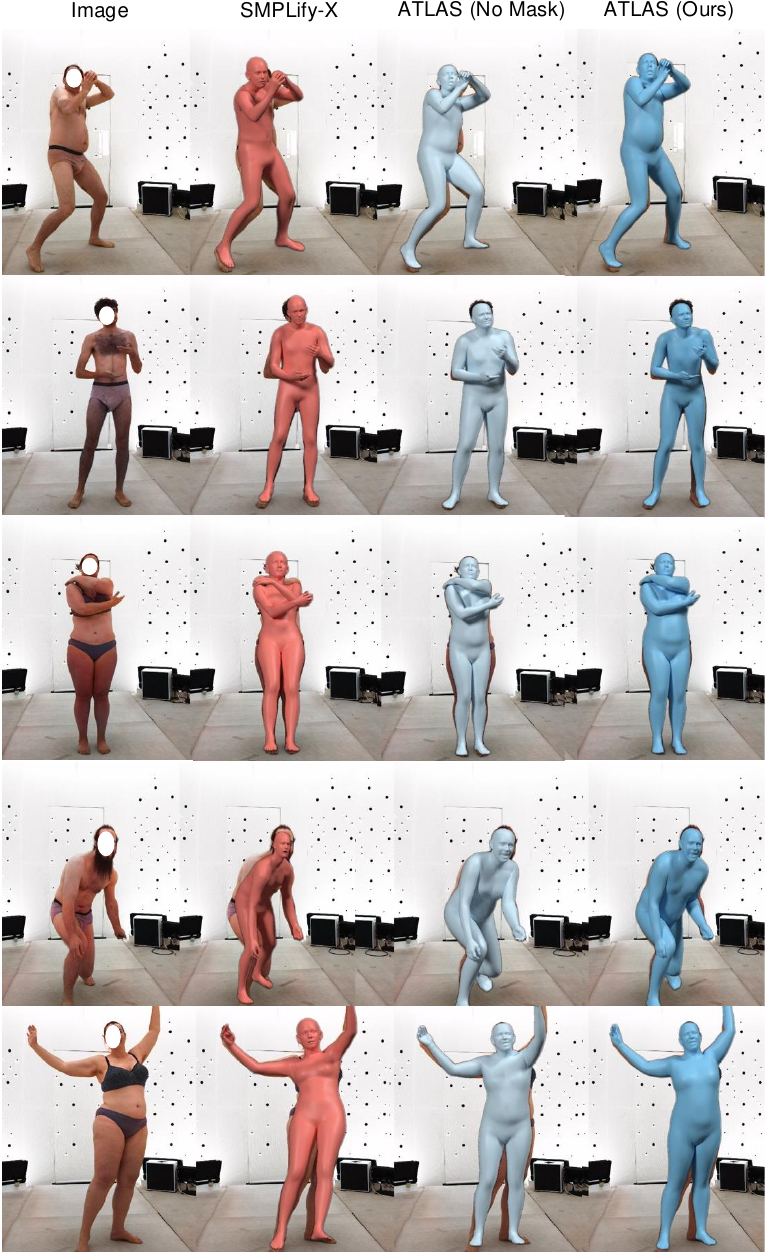}
  \vspace{-0.2in}
  \caption{\textbf{3D pose, shape, and skeleton estimation from a single image.} Our fitting pipeline captures pose more accurately, and edge gradient optimization of shape captures soft tissue attributes.}
  \label{fig:2dkps}
\end{figure}

\subsection{Monocular Mesh Fitting}
We evaluate our proposed single image mesh fitting approach (Sec.~\ref{method:fitting}) on $200$ scans from $10$ unseen subjects in the Goliath-Test dataset. Table~\ref{tab:2dkps} reports mean vertex-to-vertex error (mm) and 3D joint error (mm) after Procrustes alignment. ATLAS achieves better vertex and joint fits compared to SMPLify-X~\cite{SMPL-X:2019}, with further improvements from relative depth and mask optimization. Figure~\ref{fig:2dkps} shows that, unlike SMPLify-X, ATLAS fits pose and skeleton to keypoints without spuriously altering body shape. With edge gradient optimization, the body shape better aligns with subjects, particularly in the torso and legs. Our decoupled skeleton and shape body model, combined with decoupled keypoint and mask fitting, enables accurate pixel-aligned fitting across diverse images, as in Figure \ref{fig:shutterstock}.

\vspace{-0.1in}
\section{Conclusion}
\vspace{-0.1in}
We propose ATLAS, an expressive body model that explicitly decouples surface shape from internal skeleton. Our body model enables direct controllability of the internal skeleton, avoids spurious and incorrect correlations between surface vertices and internal joint centers, and enables decoupled skeleton and shape fitting. We additionally propose a sparse and non-linear pose corrective function that demonstrates improved generalizability for 3D human mesh modeling. Additionally, we also present a keypoint fitting framework that achieves accurate, pixel-aligned fits using ATLAS on monocular images. ATLAS represents a step toward addressing limiting assumptions in body modeling, advancing the field toward realistic, accurate, and anatomically consistent 3D human mesh modeling.

\vspace{1mm}
\noindent\textbf{Limitations.} While ATLAS captures diverse body shapes, our 15,000 subjects do not span the full range of human variation. High-resolution human scan collection and processing remains time-consuming and costly, creating a bottleneck for scaling human modeling. However, ATLAS provides an accurate prior for human scan registration, enabling development of next-generation parametric models.

{
    \small
    \bibliographystyle{ieeenat_fullname}
    \bibliography{main}
}

\clearpage

\appendix
\setcounter{figure}{11}
\setcounter{table}{3}

\section{Supplementary Overview}
In the supplementary video, we present video results of fitting ATLAS to high-fidelity 3D scans, demonstrate controllability of skeletal attributes for a dynamnic sequence, and show results of fitting ATLAS to RGB videos in the wild. 

In this supplementary document, we provide additional details on skeletal attributes, visualizations of the training data, implementation details, and qualitative results of ATLAS. The sections are organized as follows:
\begin{itemize}
    \item Section \ref{sec:skeleton_details} provides additional details regarding the 76 individually controllable skeletal attributes of ATLAS.
    \item Section \ref{sec:skinning} outlines the specific formulation of the Linear Blend Skinning (LBS) function.
    \item Section \ref{sec:internal_samples} visualizes some sample registrations from our Goliath dataset.
    \item Section \ref{sec:training_details} contains additional details regarding the training of ATLAS and the pose prior.
    \item Section \ref{sec:skin_weights} shows the skin weights before and after training.
    \item Section \ref{sec:scale_shape_space} includes visualizations of the first few external shape \& internal skeleton latent components.
    \item Section \ref{sec:full_body_sample} demonstrates the full expressiveness of ATLAS by visualizing generated subjects through random sampling of external shape, internal skeleton, body poses, hand poses, and facial expressions.
    \item Section \ref{sec:more_shutter} provides additional results on our single image to mesh prediction pipeline on in-the-wild images.
\end{itemize}
\begin{figure*}[ht]
    \centering
    \begin{tabular}{cccc}
        \begin{subfigure}[t]{0.22\textwidth}
            \centering
            \vspace{-1.5em}
            \includegraphics[width=\textwidth]{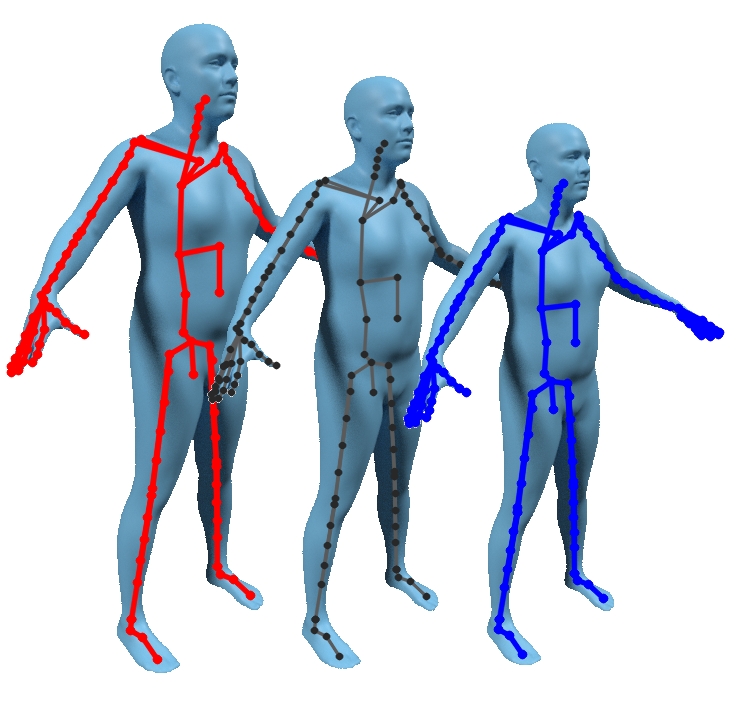}
            \vspace{-1.5em}
            \caption{Body Scale}
        \end{subfigure} &
        \begin{subfigure}[t]{0.22\textwidth}
            \centering
            \vspace{-1.5em}
            \includegraphics[width=\textwidth]{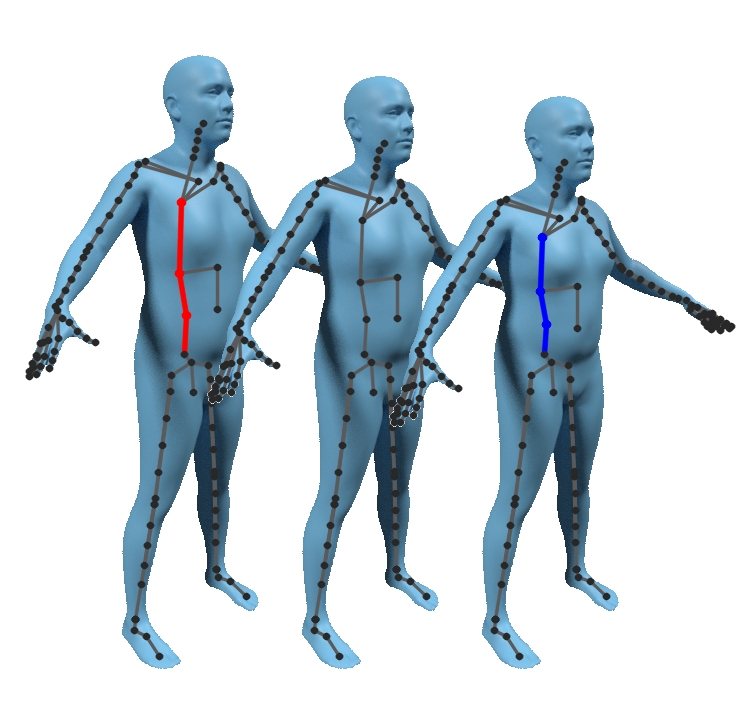}
            \vspace{-1.5em}
            \caption{Spine Length}
        \end{subfigure} &
        \begin{subfigure}[t]{0.22\textwidth}
            \centering
            \vspace{-1.5em}
            \includegraphics[width=\textwidth]{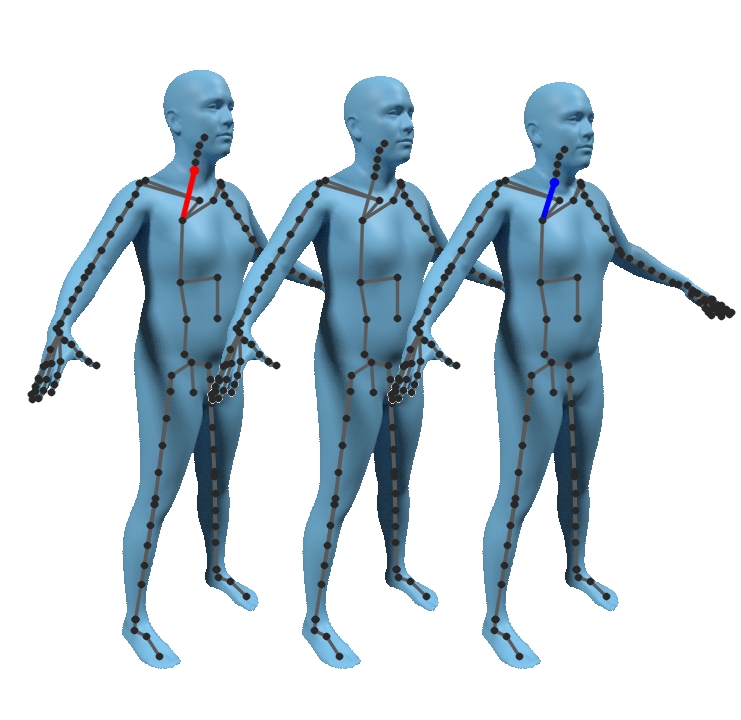}
            \vspace{-1.5em}
            \caption{Neck Offset}
        \end{subfigure} &
        \begin{subfigure}[t]{0.22\textwidth}
            \centering
            \vspace{-1.5em}
            \includegraphics[width=\textwidth]{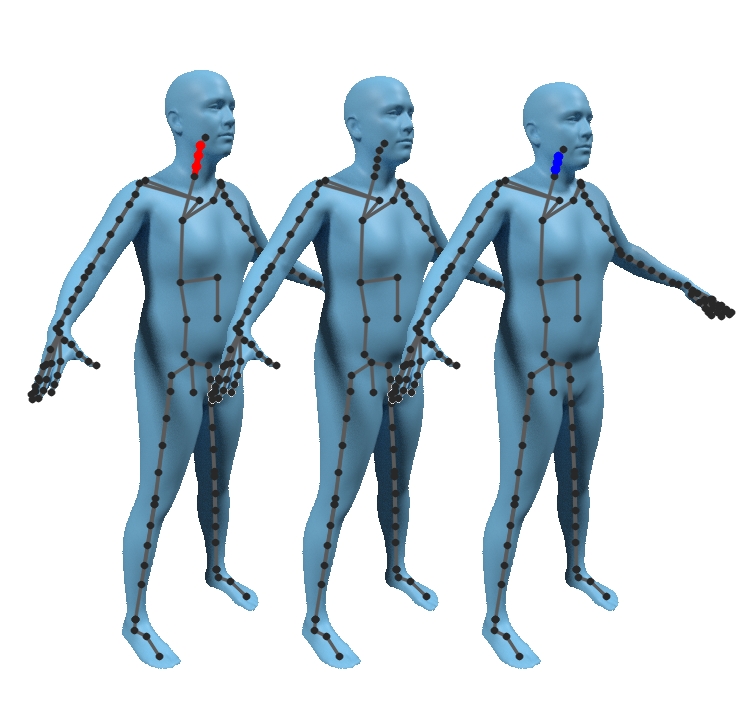}
            \vspace{-1.5em}
            \caption{Neck Length}
        \end{subfigure} \\[1em]
        
        \begin{subfigure}[t]{0.22\textwidth}
            \centering
            \vspace{-0.5em}
            \includegraphics[width=\textwidth]{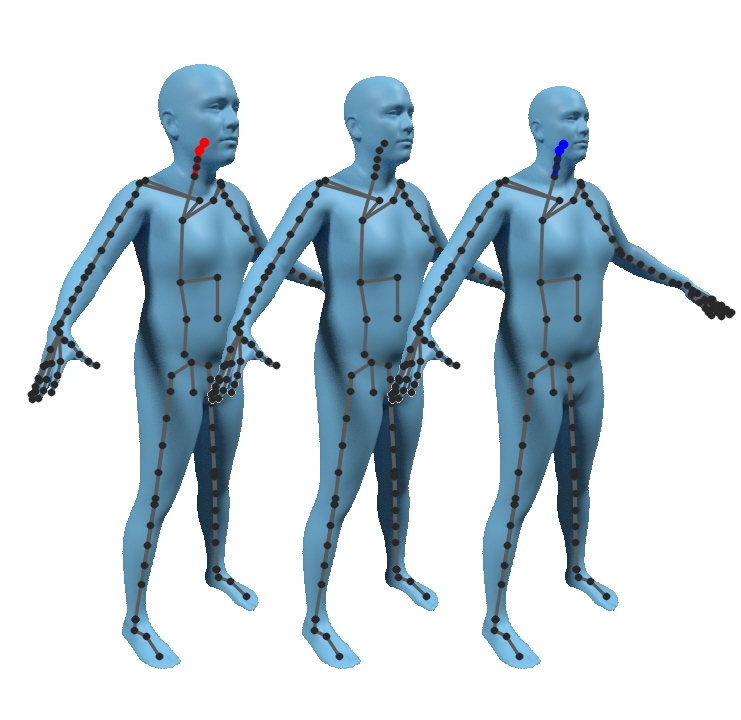}
            \vspace{-1.5em}
            \caption{Head Scale}
        \end{subfigure} &
        \begin{subfigure}[t]{0.22\textwidth}
            \centering
            \vspace{-0.5em}
            \includegraphics[width=\textwidth]{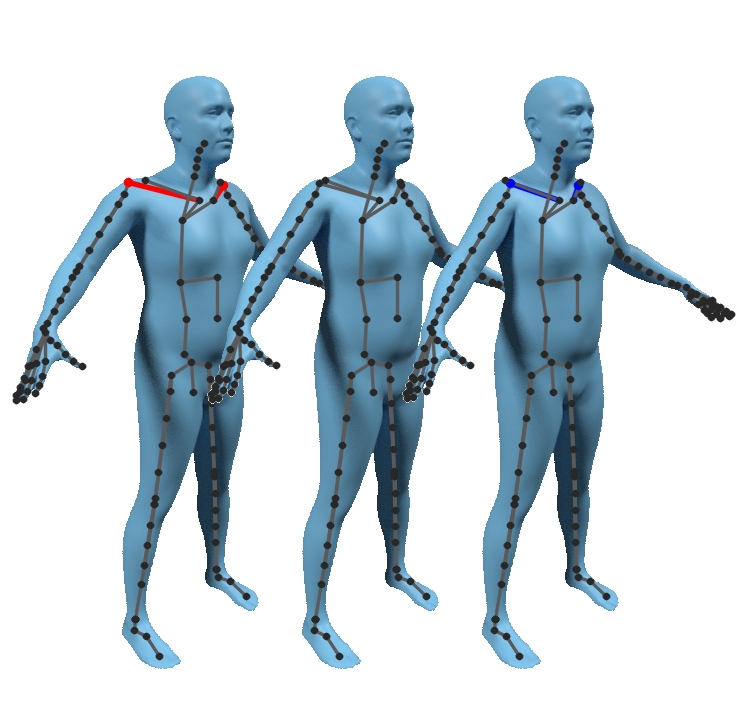}
            \vspace{-1.5em}
            \caption{Shoulder Width}\label{fig:skeletal_attributes_f}
        \end{subfigure} &
        \begin{subfigure}[t]{0.22\textwidth}
            \centering
            \vspace{-0.5em}
            \includegraphics[width=\textwidth]{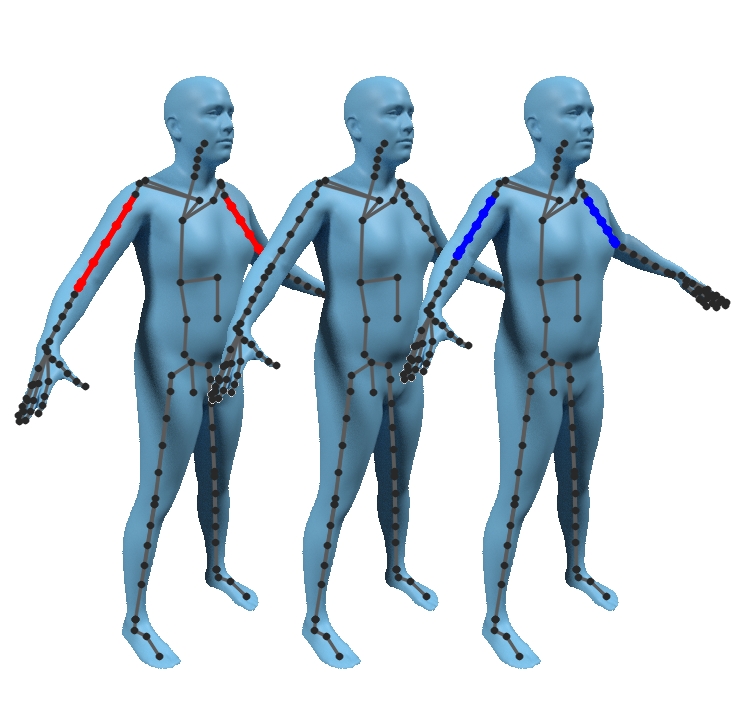}
            \vspace{-1.5em}
            \caption{Upper Arm Length}
        \end{subfigure} &
        \begin{subfigure}[t]{0.22\textwidth}
            \centering
            \vspace{-0.5em}
            \includegraphics[width=\textwidth]{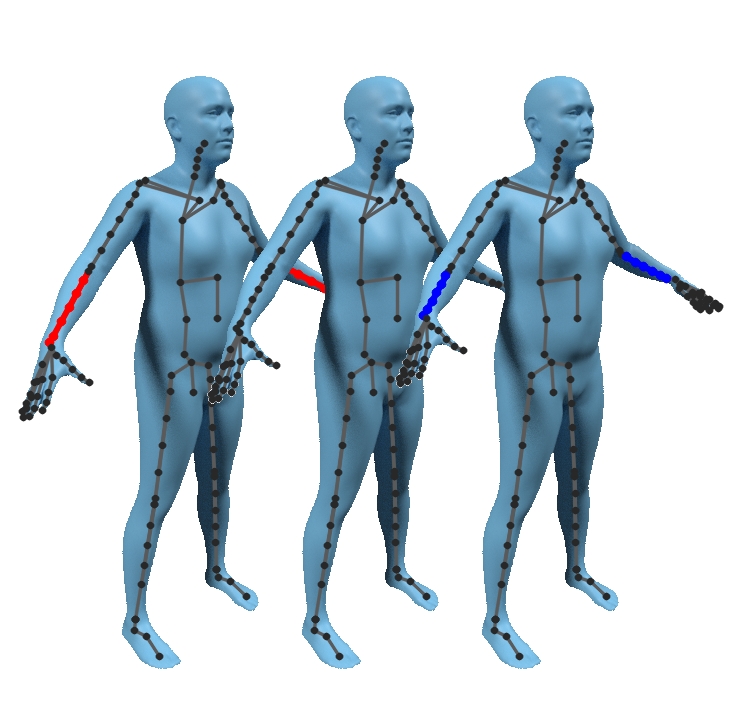}
            \vspace{-1.5em}
            \caption{Lower Arm Length}
        \end{subfigure} \\[1em]
        
        \begin{subfigure}[t]{0.22\textwidth}
            \centering
            \vspace{-0.5em}
            \includegraphics[width=\textwidth]{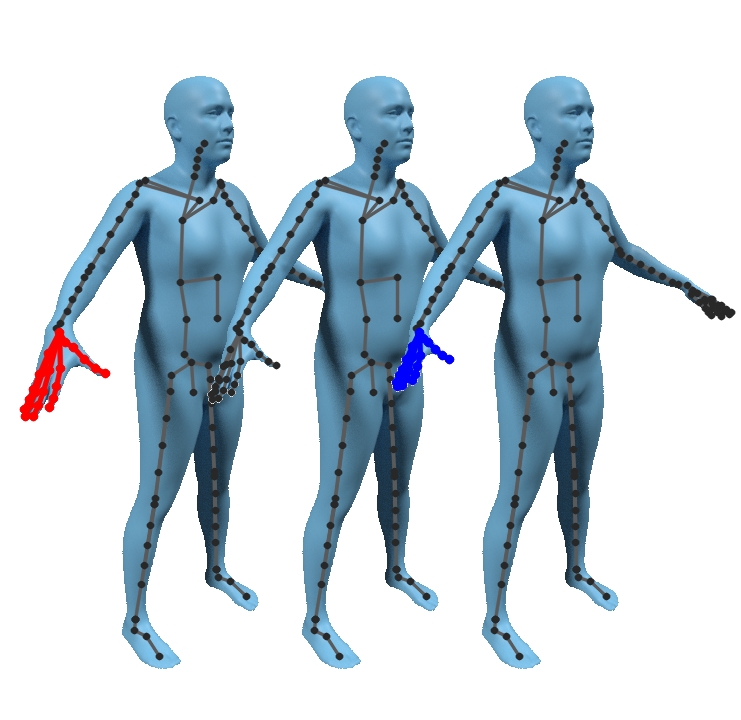}
            \vspace{-1.5em}
            \caption{Right Hand Scale}\label{fig:skeletal_attributes_i}
        \end{subfigure} &
        \begin{subfigure}[t]{0.22\textwidth}
            \centering
            \vspace{-0.5em}
            \includegraphics[width=\textwidth]{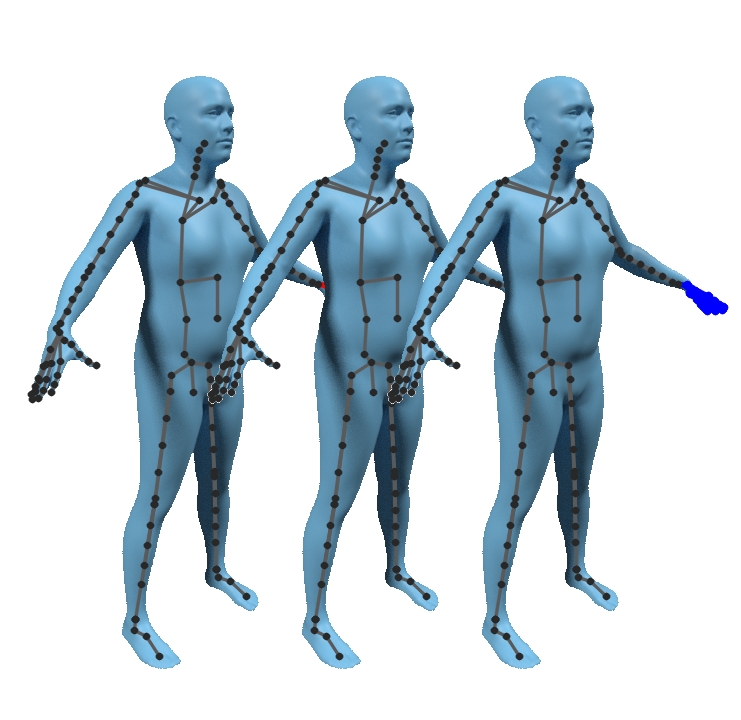}
            \vspace{-1.5em}
            \caption{Left Hand Scale}
        \end{subfigure} &
        \begin{subfigure}[t]{0.22\textwidth}
            \centering
            \vspace{-0.5em}
            \includegraphics[width=\textwidth]{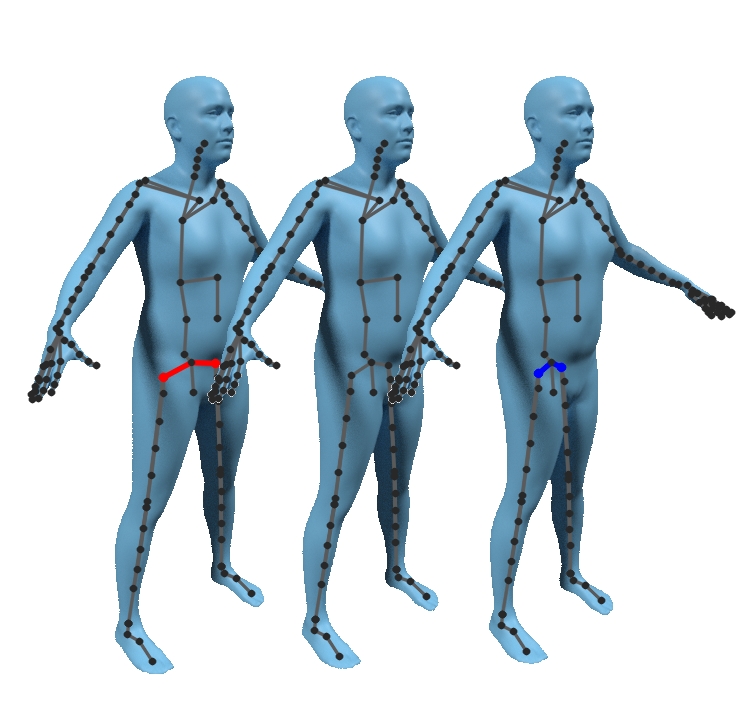}
            \vspace{-1.5em}
            \caption{Hip Width}
        \end{subfigure} &
        \begin{subfigure}[t]{0.22\textwidth}
            \centering
            \vspace{-0.5em}
            \includegraphics[width=\textwidth]{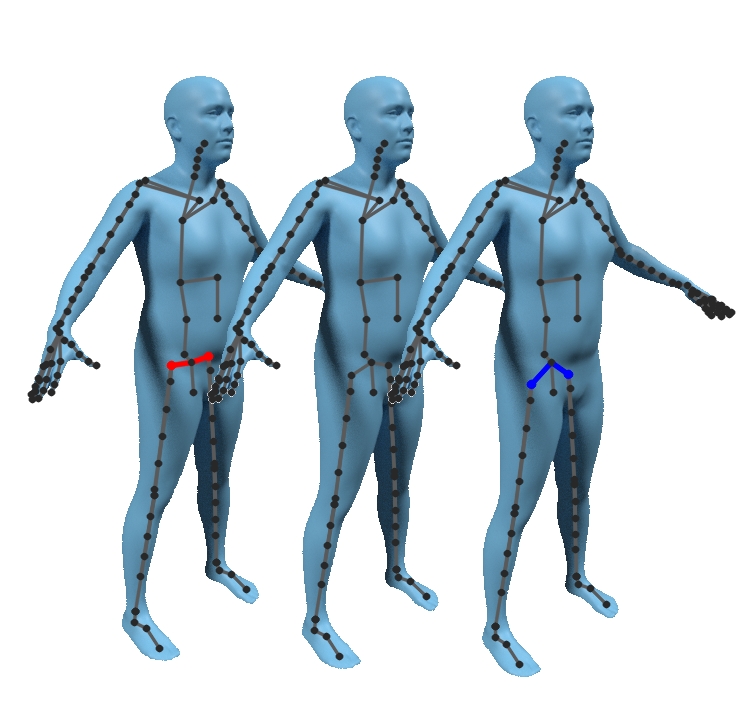}
            \vspace{-1.5em}
            \caption{Hip Height}
        \end{subfigure} \\[1em]
        
        \begin{subfigure}[t]{0.22\textwidth}
            \centering
            \vspace{-0.5em}
            \includegraphics[width=\textwidth]{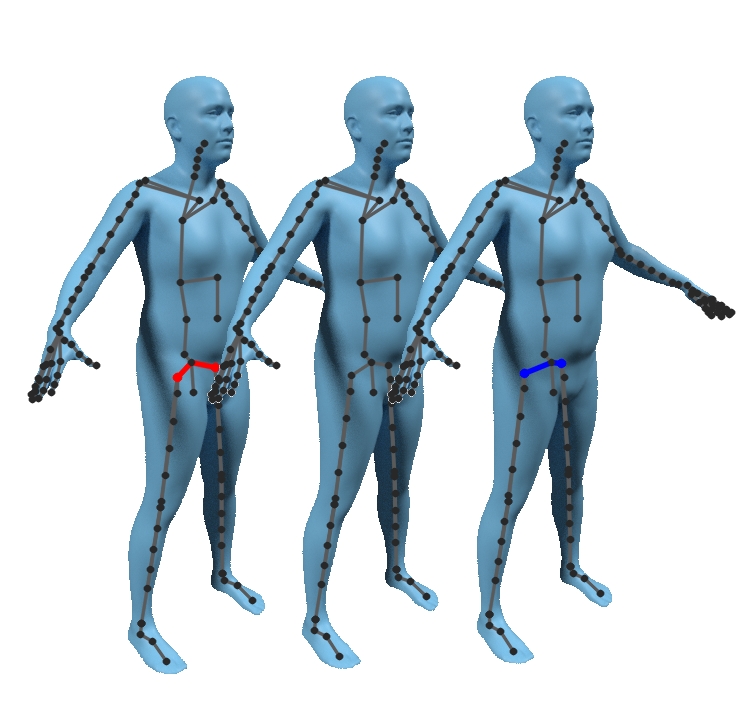}
            \vspace{-1.5em}
            \caption{Hip Depth}
        \end{subfigure} &
        \begin{subfigure}[t]{0.22\textwidth}
            \centering
            \vspace{-0.5em}
            \includegraphics[width=\textwidth]{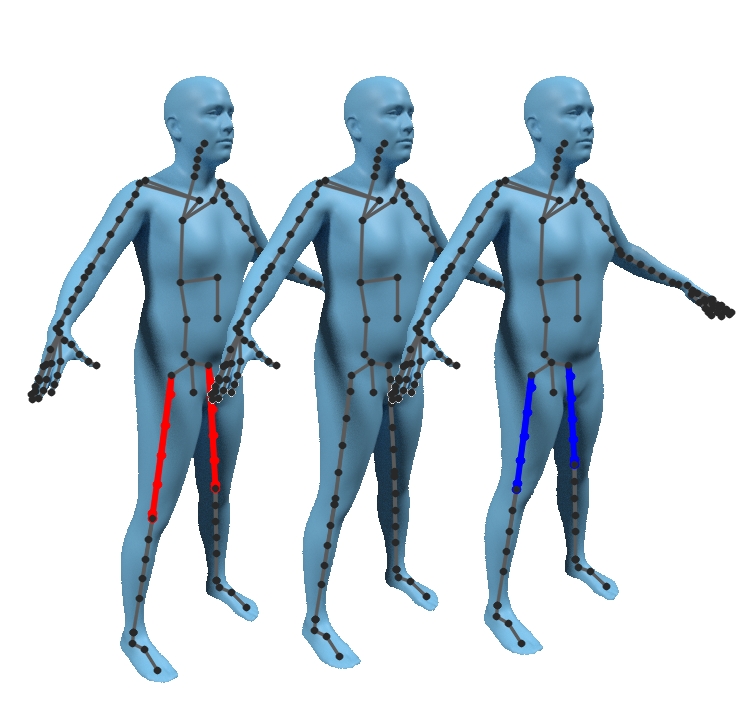}
            \vspace{-1.5em}
            \caption{Upper Leg Length}
        \end{subfigure} &
        \begin{subfigure}[t]{0.22\textwidth}
            \centering
            \vspace{-0.5em}
            \includegraphics[width=\textwidth]{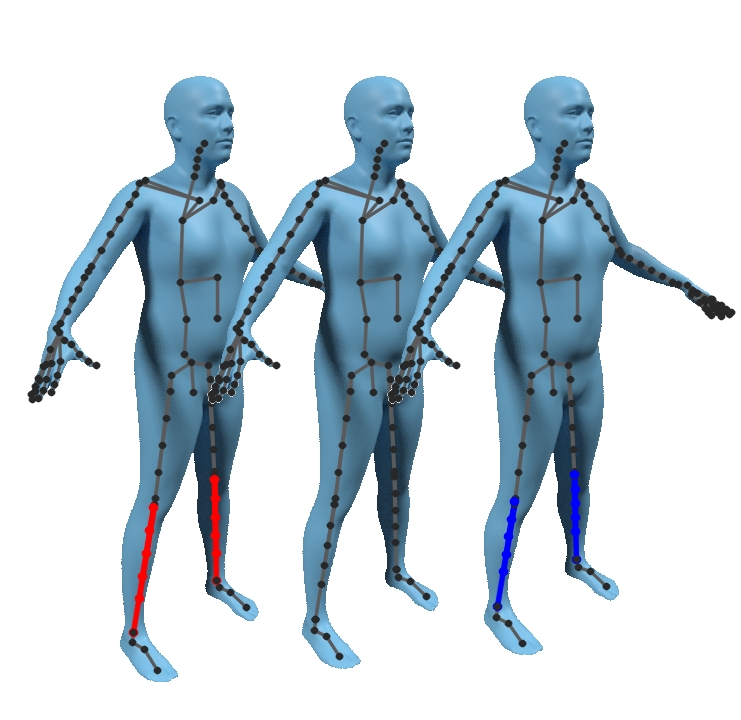}
            \vspace{-1.5em}
            \caption{Lower Leg Length}
        \end{subfigure} &
        \begin{subfigure}[t]{0.22\textwidth}
            \centering
            \vspace{-0.5em}
            \includegraphics[width=\textwidth]{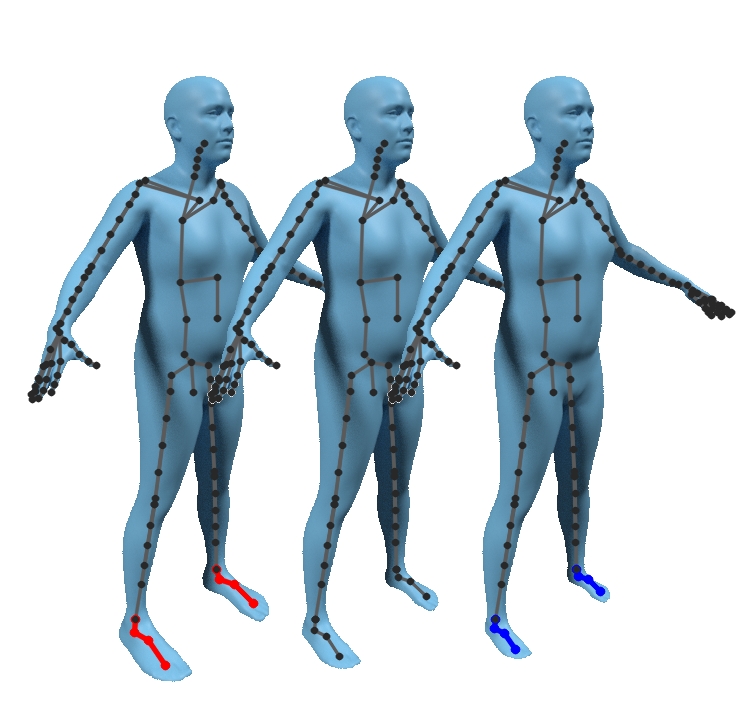}
            \vspace{-1.5em}
            \caption{Feet Scale}
        \end{subfigure} \\[1em]
    \end{tabular}
    \caption{\textbf{Visualization of Body Skeletal Attributes.} For each skeletal attribute, we show three meshes - increasing the skeletal parameter, the base mesh, and decreasing the parameter. Bones affected by the changed parameter are colored red if they have increased in size, and blue if they have decreased. Each attribute either directly scales a local joint space, including those of its kinematic children, or adjusts joint translations relative to its own kinematic parent. For instance, Figure \ref{fig:skeletal_attributes_i} shows an instance of the former, where the entirety of the right hand changes in size, while Figure \ref{fig:skeletal_attributes_f} is an instance of the latter, where the shoulder joint center is moved, driving an increase or decrease in shoulder width.}
    \vspace{-1em}
    \label{fig:skeleton_details}
\end{figure*}

\begin{table}[b]
\small
\begin{center}
    \renewcommand{\arraystretch}{0.9}
    \begin{tabular}{cccc}
    \toprule
    \textbf{Shape} & \textbf{Skeleton} & \textbf{3DBodyTex} & \textbf{Goliath-Test} \\
    \midrule
    \cmark & \xmark & 6.47 & 4.76 \\
    \xmark & \cmark & 3.17 & 2.67 \\
    \cmark & \cmark & \textbf{2.48} & \textbf{2.34} \\
    \bottomrule
    \end{tabular}
\vspace{-0.1in}
\caption{Mesh fitting error (mm) with shape and skeleton params.}
\label{table:decouple}
\end{center}
\end{table}

\section{Details on Controllable Skeletal Attributes}\label{sec:skeleton_details}
ATLAS defines 76 controllable skeletal attributes that modify different parts of the skeleton. As described in the main paper, 15 of these attributes directly scale a local joint space (and those of its kinematic children). These consist of scales that affect the full-body, head, hands, feet, and individual fingers. The remaining 61 are bone length parameters that directly adjust each joint's center location with respect to its kinematic parent. These include the spine, neck offset, neck length, shoulder width, upper arms, lower arms, hip location, upper legs, lower legs, and each bone in the finger for precise controllability. We visualize the skeletal attributes that affect major parts (excluding individual finger bone adjustments) in Figure \ref{fig:skeleton_details}. 

Further, we demonstrate that the surface shape basis and the skeletal basis are both necessary and are complementary by evaluating mesh fitting with disentangled parameters in Table \ref{table:decouple}. Shape alone misses height and limb length variations, while skeleton alone overlooks soft tissue. Using both, like ATLAS, best captures diverse body shapes.

\section{Linear Blend Skinning Formulation}\label{sec:skinning}
In this section, we provide the precise formulation for the LBS skinning function $M$ used in Section \textcolor{iccvblue}{3.1} of the main paper. This transformation $M$ to yield a scaled and posed vertex $x_i$ is written as:
\begin{equation}
    x_i = \sum_{j=1}^I \omega_{ij} \mathcal{T}_j(\bar{\theta}, \bar{t}, \theta, \sigma, t)\mathcal{T}_j(\bar{\theta}, \bar{t}, \vec{0}, \vec{0}, \vec{0})^{-1}\tilde{x}_i
\end{equation}
where $\bar{\theta}$ and $\bar{t}$ define the rest pose of the skeleton. These rest pose definitions of each joint's rotation and offset with respect to its parent are necessary because unlike SMPL~\cite{SMPL:2015} where each joint's coordinate system is root axis-aligned, our rotations are skeleton-aligned. The forward kinematic transformation $\mathcal{T}_j$ is then defined by:
\begin{small}
\begin{equation}
    \mathcal{T}_j(\bar{\theta}, \bar{t}, \theta, \sigma, t) = \Pi_{a\in\boldsymbol{K}(j)} 
    \begin{bmatrix}
        2^{\sigma(a)}R(\theta_a)R(\bar{\theta}_a) & t(a)t_e(a) + \bar{t}_l \\
        0 & 1
    \end{bmatrix}
\end{equation}
\end{small}
where $\boldsymbol{K}(j)$ are the kinematic tree parents of joint $a$ in ascending order, $\sigma(a)$, $t(a)$, and $t_e(a)$ are zero if joint $a$ lacks a corresponding skeleton modification. Thus, $\mathcal{T}_j(\bar{\theta}, \bar{t}, \vec{0}, \vec{0}, \vec{0})^{-1}$ transforms from global to joint-$j$'s local coordinates through kinematic tree traversal of an unposed, unscaled skeleton, while $\mathcal{T}_j(\bar{\theta}, \bar{t}, \theta, \sigma, t)$ transforms from joint-$j$'s local to global coordinates with skeleton posing and bone scale/length modifications.

\section{Visualization of scans from the Goliath dataset}\label{sec:internal_samples}
\begin{figure*}[ht]
  \centering
  \includegraphics[width=\linewidth]{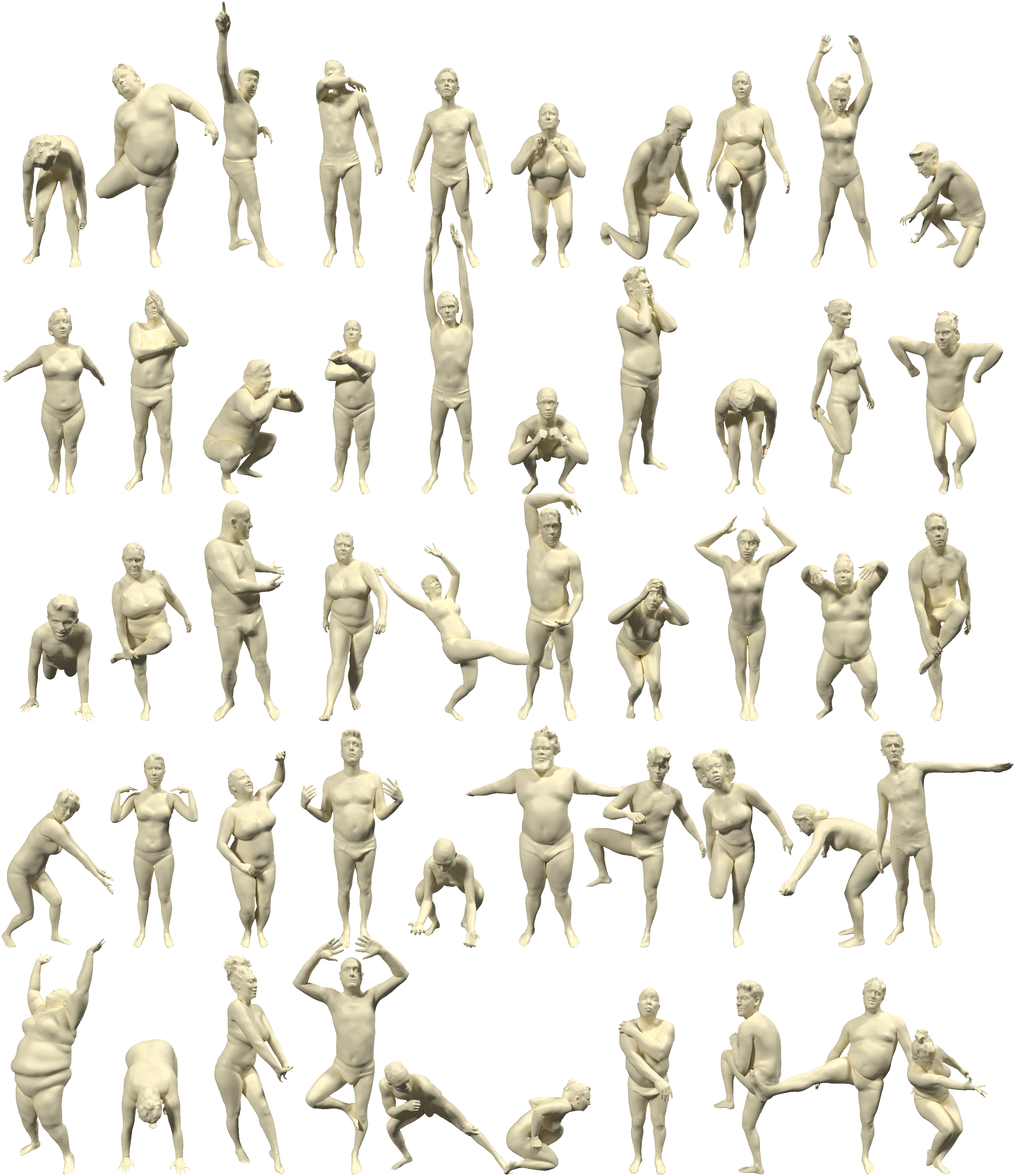}
  \caption{\textbf{Sampled Visualizations of Our Multi-Pose Dataset.} We train ATLAS on a diverse set of 600k scans captured by a high-resolution scanner with 240 synchronized cameras.}
  \label{fig:internal_samples}
\end{figure*}

In Figure \ref{fig:internal_samples} we provide a sample of our Goliath dataset To assemble a large and diverse set of of scans to train our model, we capture 130 subjects in a diverse suite of poses including conversational settings, charades acting, and dynamic movements. The frames are captured using 240 high-resolution, synchronized cameras that yields meshes with approximately 1 million vertices. The scans are captured at 30-90 FPS, and we use furthest-point-sampling on pose to select an interesting and diverse set of 600k frames to train ATLAS.

\begin{figure*}[t]
  \centering
  \includegraphics[width=0.8\linewidth]{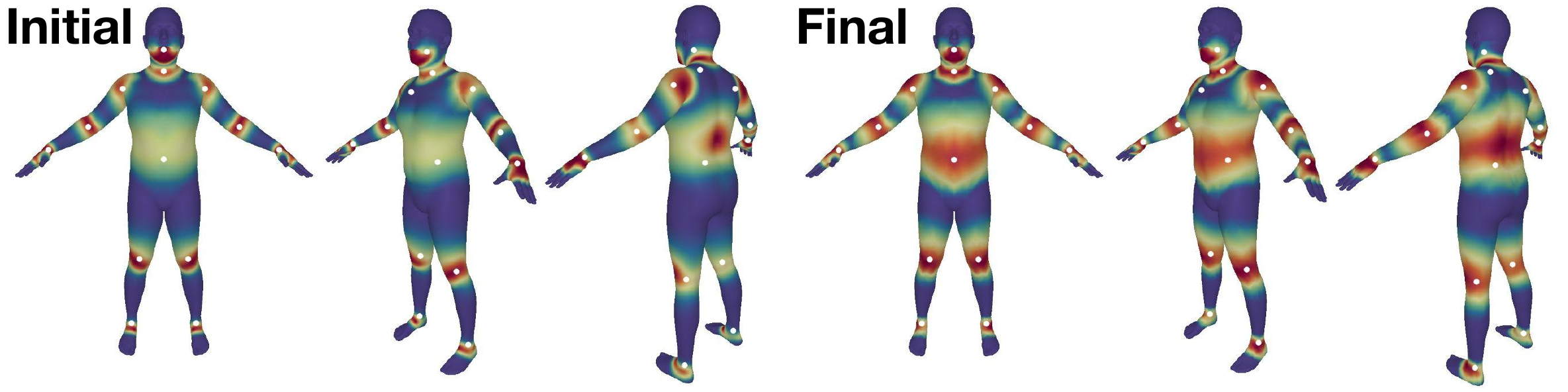}
  \caption{Skinning weights for the jaw, neck, upper arm, elbow, wrist, lower spine, knee, and ankle before and after optimization.}
  \label{fig:rebut_skin_weights}
\end{figure*}
\section{Training Details}\label{sec:training_details}
\subsection{ATLAS Body Model Details}
\subsubsection{Vertex Resolutions}
While ATLAS is natively trained at the highest resolution with 115,834 vertices, we define mappings to the $6890$ and $10475$ vertices of SMPL and SMPL-X. This enables transformations between ATLAS and SMPL/SMPL-X and allows ATLAS to operate with fewer vertices for improved efficiency.

\subsubsection{Body Model Design}
ATLAS leverages a joint structure designed by expert sculpting artists to ensure anatomical consistency. The joint locations adhere to the human bone structure, and in place of a standard single 3DoF rotation for major joints, ATLAS decomposes them into anatomically accurate sub-joints. For example, the shoulder includes a scapular joint, and the ankle is divided into subtalar and talocrural joints. 

\subsection{ATLAS Training Details}
ATLAS is trained end-to-end by sampling registrations with their corresponding rest-pose surface vertices, internal skeletal parameters, and full body pose. The surface vertices and skeletal parameters are input into their respective linear autoencoders~\cite{ng2011sparse}, the pose is input into our sparse, non-linear pose correctives function, and the mesh is rigged with the reconstructed vertices, reconstructed skeletal parameters, pose correctives, and trainable skin weights. 

We initialize autoencoders~\cite{ng2011sparse} using PCA of surface vertices and skeletal parameters from our multi-shape dataset. For each training iteration, we sample the number of components $n \in [1, \text{max}]$ and preserve only the first $n$ features in the autoencoder latent bottleneck, zeroing out the remainder. This ordered dropout strategy maintains the component importance hierarchy throughout optimization. We use 128 components for the shape and 16 components for the skeleton, as we found fitting error plateaued beyond these components.

ATLAS is trained by minimizing the loss:
\begin{align*}
    \mathcal{L} &= \mathcal{L}_{\text{data}} + \mathcal{L}_{\text{shape\_reg}} + \mathcal{L}_{\text{skele\_reg}} + \mathcal{L}_{\text{skin\_lapl}} + \mathcal{L}_{\text{pc\_lapl}} \\
    &+ \mathcal{L}_{\text{skin\_init}} + \mathcal{L}_{\text{pc\_act\_reg}}
\end{align*}
where $\mathcal{L}_{\text{data}}$ is the main data term minimizing vertex-to-vertex distance between the registration and the predicted mesh. $\mathcal{L}_{\text{shape\_reg}}$ and $\mathcal{L}_{\text{skele\_reg}}$ are L2 losses that regularize the intermediate latents of the surface vertex and skeleton attribute autoencoders. $\mathcal{L}_{\text{skin\_lapl}}$ and $\mathcal{L}_{\text{pc\_lapl}}$ regularize the skin weights and pose corrective blendshapes with a cotangent laplacian loss. $\mathcal{L}_{\text{skin\_init}}$ regularizes the skin weights towards their artist-defined initialization through L2. $\mathcal{L}_{\text{pc\_act\_reg}}$ imposes an L1 regularization loss on the pose corrective activation matrix, which is geodesic initialized, to encourage sparsity in vertex-joint correlations. 

\subsection{Pose Prior Implementation Details}
For our pose prior, we adopt a lightweight VAE architecture similar to that of SMPL-X \cite{SMPL-X:2019}. The VAE has a 32 latent dimension, takes as input 6D continuous rotation vectors for the full body excluding hands, and is trained to reconstruct samples from our 600k multi-pose dataset. The model is trained for 40 epochs with a batch size of 512 and a learning rate of 5e-3. We minimize three losses - the KL divergence loss, a reconstruction loss, and the angle difference loss between the input and output.

\section{Optimized Skin Weights}\label{sec:skin_weights}
We initialize skin weights $\Omega$ with artist-defined values and optimize them end-to-end during training. The weights before and after training are shown in Figure \ref{fig:rebut_skin_weights}.

\begin{figure*}[h]
  \centering
  \includegraphics[width=\linewidth]{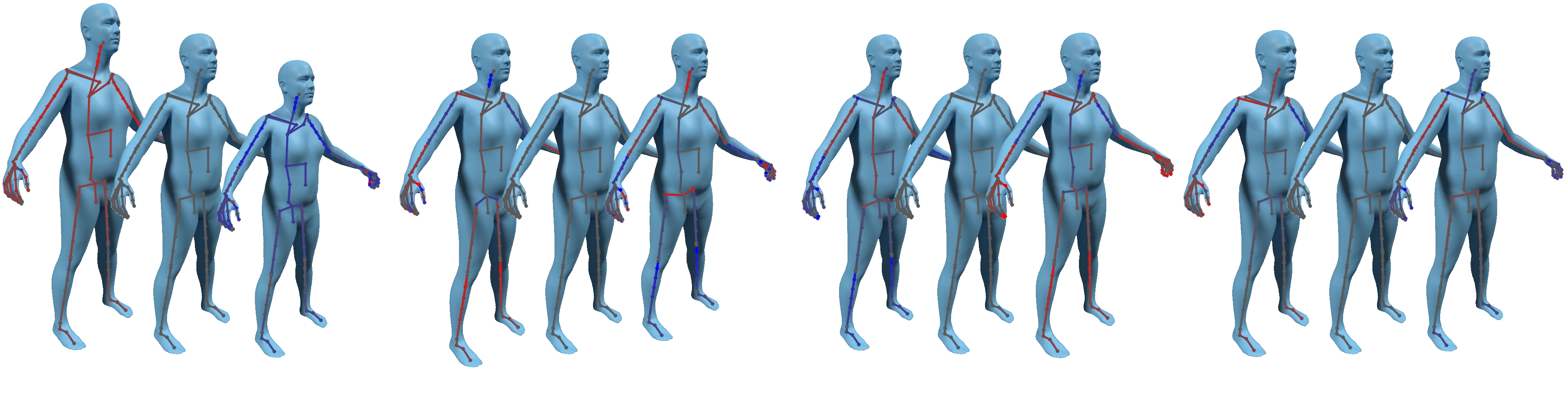}
  \caption{\textbf{Visualizations of the first four internal skeleton components.} For each component, we visualize changes in the mesh from decreasing and increasing the component. The skeleton is colored such that red indicates an increase in bone length while blue indicates a decrease. The skeletal components alone are sufficient to capture most human body variation. The first component is correlated with overall size of the subject, the second captures the neck and the hips, the third focuses on the shoulders and arms (decoupling upper and lower arm lengths), while the fourth captures length of the full arm.}
  \label{fig:scale_latent}
\end{figure*}

\begin{figure*}[h]
  \centering
  \includegraphics[width=\linewidth]{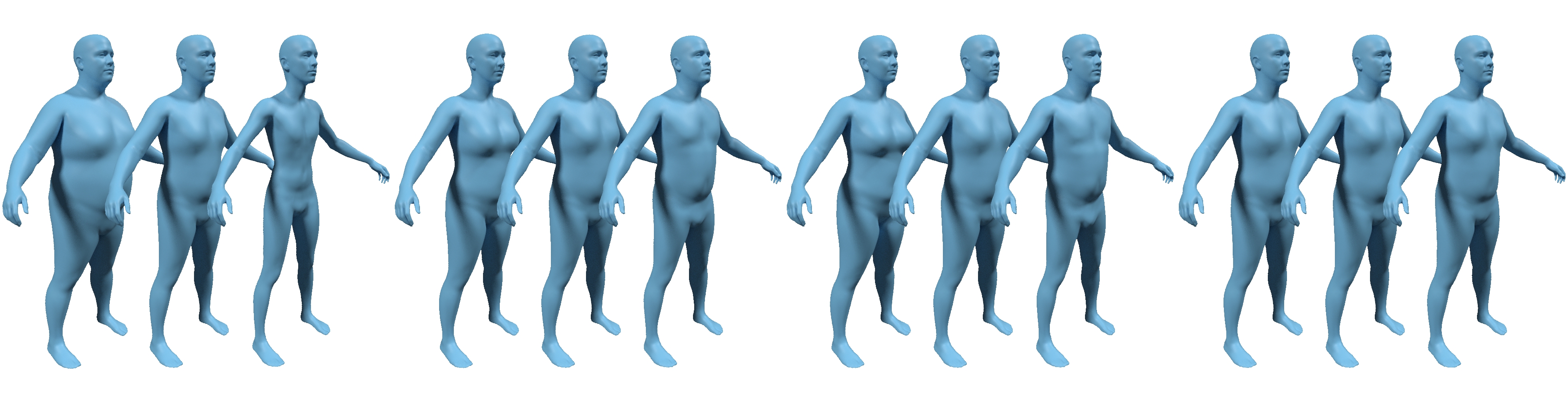}
  \caption{\textbf{Visualizations of the first four external surface components.} As most of the body variation (height, arm length, hand size, etc) are already captured by the skeleton, the surface components focus on soft tissue changes such as weight, neck width, arm thickness, and facial attributes. Note that we do not display the skeleton as it remains unchanged with variations in the surface vertices.}
  \label{fig:shape_latent}
\end{figure*}

\begin{figure*}[ht]
  \centering
  \includegraphics[width=0.92\linewidth]{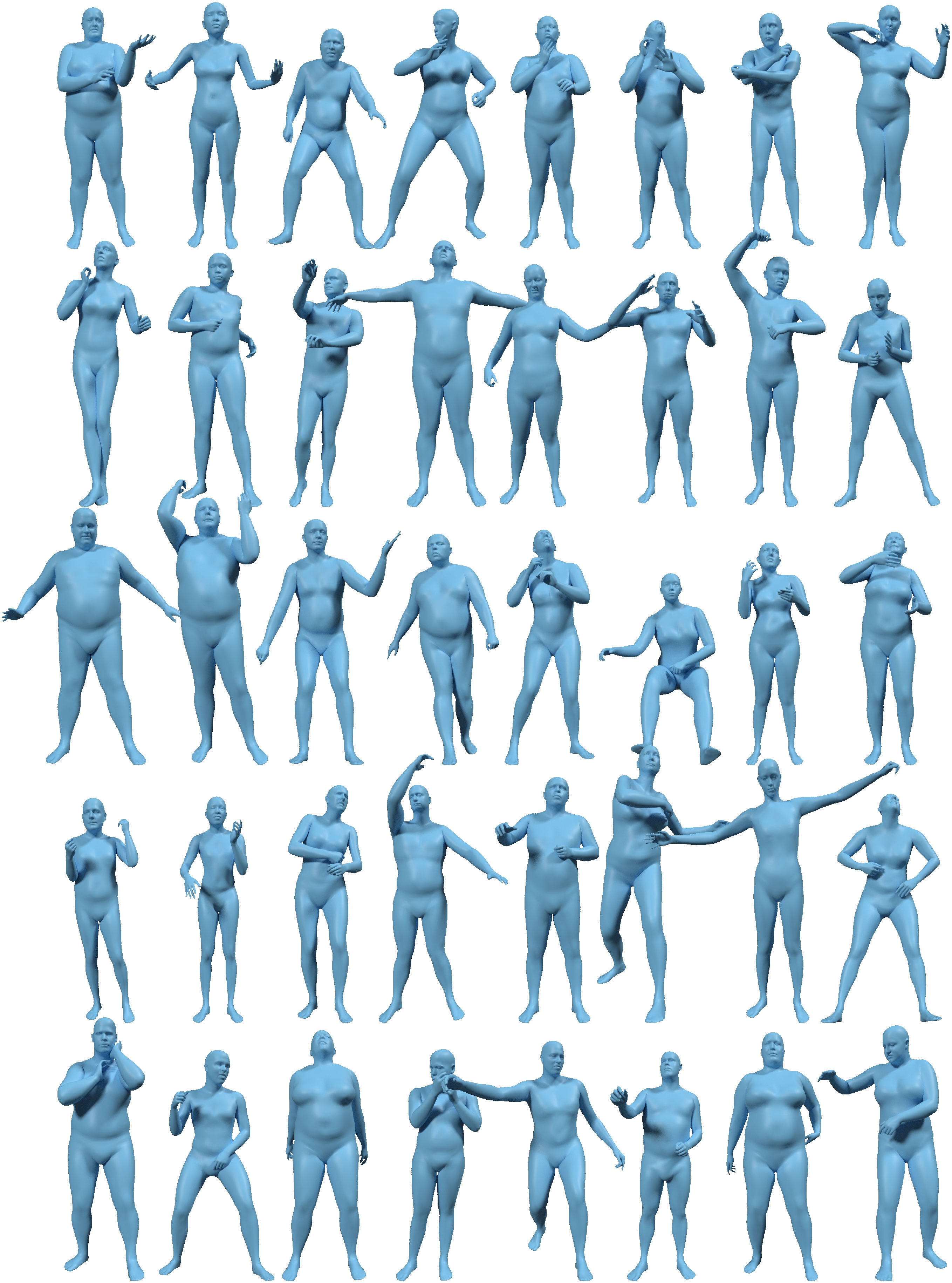}
  \caption{\textbf{Visualization of Random Latent Samples from ATLAS.} We randomly sample subject surface vertices and internal skeleton from the latent spaces, sample pose from our VAE pose prior and hand PCA space, and facial expressions from the FLAME space. ATLAS captures a wide breadth of realistic human shapes and articulates them into realistic poses.}
  \label{fig:random_samples}
\end{figure*}

\begin{figure*}[ht]
  \centering
  \includegraphics[width=\linewidth]{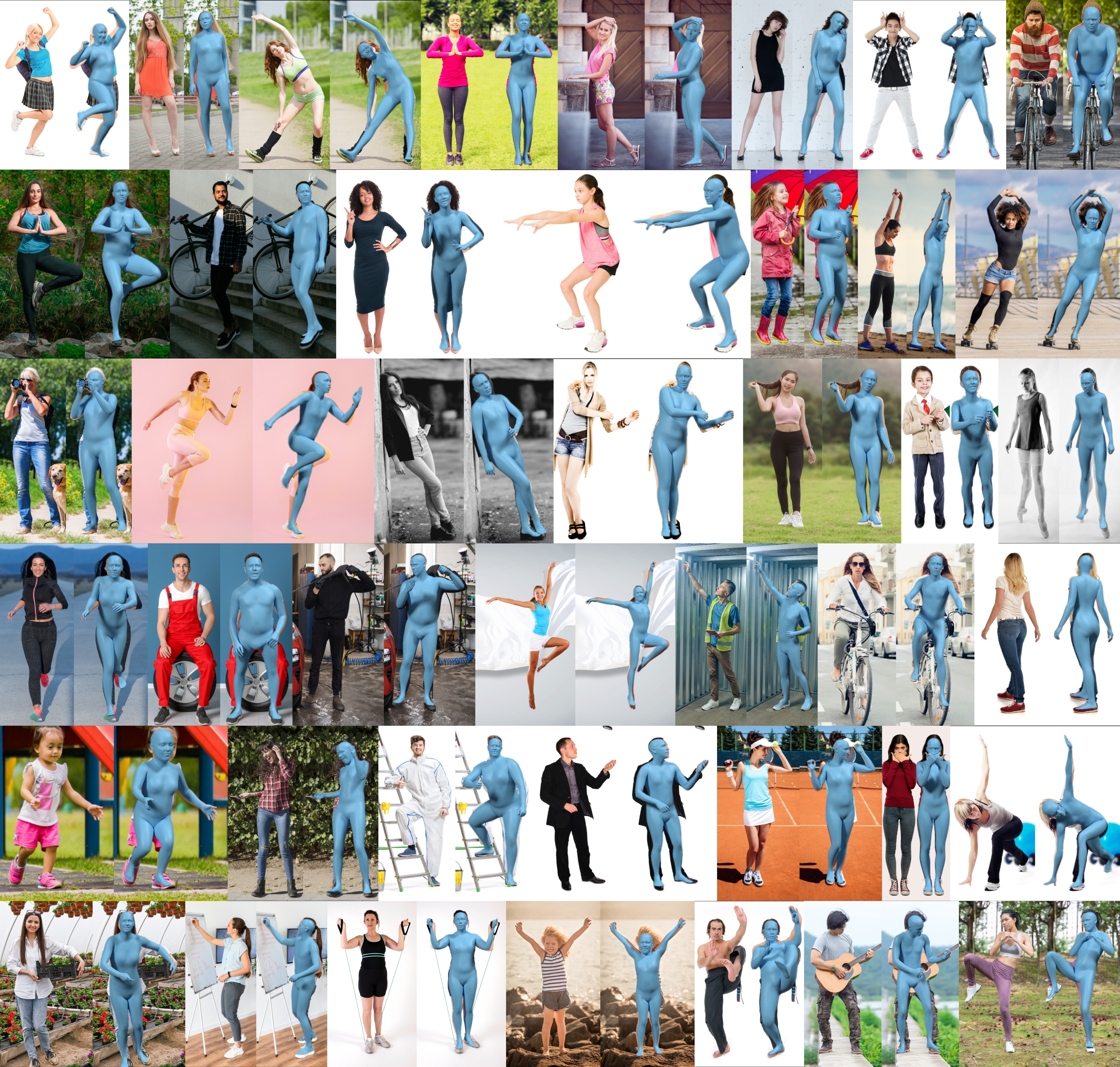}
  \caption{\textbf{Additional Visualizations of Fitting ATLAS to Single Images.} Our fitting pipeline can capture a wide range of poses and shapes in addition to facial expressions.}
  \label{fig:shutter_more}
\end{figure*}

\section{Skeleton and Shape Latent Spaces}\label{sec:scale_shape_space}
Our skeletal attribute definitions allow for direct controllability of individual aspects of the internal skeleton. Furthermore, for lower-dimensional keypoint fitting, scan registration, and skeleton modification, our skeleton latent space provides data-driven correlations between different aspects of the body. We visualize the first four components of the skeletal components in Figure \ref{fig:scale_latent}. We find that the skeletal attributes themselves capture most of the variation in the human body, such as overall body size, shoulder width, arm length, etc. While the skeletal components focus on the internal structure of humans, the surface components, shown in Figure \ref{fig:shape_latent} instead focus on the external soft tissue changes. Our surface components are more subtle than the shape components of prior work, as previous methods entangle skeletal and surface attributes, forcing the same components to capture variations in both soft tissue attributes and internal skeleton.

\section{Latent Sampling of Shape, Skeleton, Pose, and Expressions}\label{sec:full_body_sample}
In this section, we further demonstrate the expressiveness of ATLAS by randomly sampling shaped, articulated human subjects in Figure \ref{fig:random_samples}. More specifically, we sample from our surface and skeletal latent spaces to model a random identity, then sample from our pose prior and hand PCA space for full-body pose, and finally sample facial expressions. The resulting meshes are realistic, and they span a wide range of diverse human subjects in a variety of poses.

\section{Additional Results on Mesh Prediction in the Wild}\label{sec:more_shutter}
We extend the results in Figure \textcolor{iccvblue}{11} of the main paper by providing additional results on in-the-wild images in Figure \ref{fig:shutter_more}. Our fitting procedure complements ATLAS by yielding shape, scale, pose, and expression parameters from 2D RGB images in the wild. Of particular note is ATLAS's ease at capturing undersized subjects such as children. By explicitly modeling the size of each skeletal part, ATLAS naturally predicts realistic shapes for children, accounting for their relatively larger heads compared to the rest of their body.

\end{document}